\documentclass[times,twocolumn,final,authoryear]{elsarticle}

\usepackage{ycviu}
\usepackage{framed,multirow}
\usepackage{float}

\usepackage{amssymb}
\usepackage{latexsym}

\usepackage{url}
\usepackage{xcolor}
\definecolor{newcolor}{rgb}{.8,.349,.1}

\usepackage{times}
\usepackage{epsfig}
\usepackage{graphicx}
\usepackage{amsmath}
\usepackage{amssymb}
\usepackage{subcaption}
\usepackage{mathabx}
\usepackage{epsfig}
\usepackage{times}
\usepackage{multirow}
\usepackage{hhline}
\usepackage{graphics}
\usepackage{comment}
\usepackage{anyfontsize}
\usepackage[ruled,vlined]{algorithm2e}
\usepackage{algorithmic}
\usepackage{url}
\newcommand{\etal}{\textit{et al}.}
\SetKwInput{KwInput}{Input}
\SetKwInput{KwPre}{Pre-train}

\journal{Computer Vision and Image Understanding}

\begin{document}

\ifpreprint
  \setcounter{page}{1}
\else
  \setcounter{page}{1}
\fi

\begin{frontmatter}

\title{Task dependent Deep LDA pruning of neural networks}

\author[1]{Qing \snm{Tian}\corref{cor1}} 
\cortext[cor1]{Corresponding author: 
  Tel.: +0-000-000-0000;  
  fax: +0-000-000-0000;}
\ead{qing.tian@mail.mcgill.ca}
\author[1]{Tal \snm{Arbel}}
\author[1]{James J. \snm{Clark}}

\address[1]{Centre for Intelligent Machines \& ECE Department, McGill University, \\3480 University Street, Montreal, Quebec H3A 0E9, Canada}

\received{1 May 2013}
\finalform{10 May 2013}
\accepted{13 May 2013}
\availableonline{15 May 2013}
\communicated{S. Sarkar}

\begin{abstract}
With deep learning's success, a limited number of popular deep nets have been widely adopted for various vision tasks. However, this usually results in unnecessarily high complexities and possibly many features of low task utility. In this paper, we address this problem by introducing a task-dependent deep pruning framework based on Fisher's Linear Discriminant Analysis (LDA). The approach can be applied to convolutional, fully-connected, and module-based deep network structures, in all cases leveraging the high decorrelation of neuron motifs found in the pre-decision space and cross-layer deconv dependency.
Moreover, we examine our approach's potential in network architecture search for specific tasks and analyze the influence of our pruning on model robustness to noises and adversarial attacks.
Experimental results on datasets of generic objects (ImageNet, CIFAR100) as well as domain specific tasks (Adience, and LFWA) illustrate our framework's superior performance over state-of-the-art pruning approaches and fixed compact nets (e.g. SqueezeNet, MobileNet). The proposed method successfully maintains comparable accuracies even after discarding most parameters (98\%-99\% for VGG16, up to 82\% for the already compact InceptionNet) and with significant FLOP reductions (83\% for VGG16, up to 64\% for InceptionNet). Through pruning, we can also derive smaller, but more accurate and more robust models suitable for the task.
\end{abstract}

\begin{keyword}
\MSC 41A05\sep 41A10\sep 65D05\sep 65D17
\KWD deep neural networks pruning\sep deep linear discriminant analysis\sep deep feature learning

\end{keyword}

\end{frontmatter}


\section{Introduction} \label{sec:intro}
In this paper, we explore the premise that less useful features (including their possible redundancies) in overparameterized deep nets can be pruned away to boost efficiency and accuracy. In our opinion, optimal deep features should be task-dependent. Prior to deep learning, features were usually hand-engineered with domain specific knowledge~\citep{lowe1999,ojala1996comparative,ahonen2004,kumar2009,vstruc2009}. With the success of deep learning, we no longer need to handcraft features, but people are still handcrafting various architectures, which impacts both the quality and quantity of features to be learned. Some features learned via arbitrary architectures may be of little utility for the current task at hand. Such features not only add to the storage and computational burden but may also skew the data analysis (e.g. image classification in this paper) or result in over-fitting when data is limited.
Furthermore, many researchers hand-design deep architectures on a particular large benchmark dataset (e.g. ImageNet), but expect to achieve an overall generalization ability. It is possible that such architectures hand-designed on one dataset cannot produce optimal features for other tasks, despite the large enough capacity.

Instead of handcrafting fixed nets and assuming their generalizability to various tasks, in this paper, we attempt to derive a range of deep models well-suited to the current task through task dependent network pruning (feature selection). We develop a deep Linear Discriminant Analysis (LDA)~\citep{fisher1936} based neuron/filter pruning framework that is aware of both class separation and holistic cross-layer dependency. The pruning approach strategically selects useful deep features from a discriminative dimension reduction perspective. Since possible harmful dimensions can interfere or skew the classification, our pruning approach has a potential to help with accuracy aside from efficiency gains.
Key contributions that distinguish our approach from previous ones
include: (1) Our pruning has a deep LDA neuron utility measure that is derived from final task-dependent class separation. The LDA-based utility is calculated, unraveled, and traced backwards from the final latent space where the linear assumption of LDA is reasonable and variances are more disentangled~\citep{bengio2013}. Those two factors make our LDA-based pruning directly along neuron dimensions well-grounded, which we will show in Sec.~\ref{sec:geig} through solving a generalized eigenvalue problem. In contrast, most previous pruning approaches have hard-coded or human-injected utility measures (e.g. magnitudes of weights, variances, activation) and reduce model complexity along a direction that is not necessarily desirable for the task.
(2) Through deep discriminant analysis, the proposed approach determines how many filters, and of what types, are appropriate in a given layer. By pruning deep modules, it provides a top-down strategy for architecture search. This task-utility-aware deep dimension reduction is different from a wide range of popular compact structures that employ $k$ random $1\times1$ filter sets to arbitrarily reduce feature dimension to size $k$. A small $k$ may cut the information flow to higher layers, while a large $k$ may lead to redundancy, overfitting, and interference. Such arbitrariness also exists for other filter types.
(3) Through pruning large networks of high memorization capability, the proposed method helps over-parameterized nets forget about task-unrelated factors and derive a feature subspace that is more invariant and robust to irrelevant lighting, background, noises, and so on. We also analyze the effect of our pruning on model robustness against adversarial attacks and noises. At the time of writing, few if any works have investigated such aspects in the existing literature on deep net pruning.
(4) The approach presented here handles a wide variety of structures such as convolutional, fully-connected, modular, and hybrid ones and prunes a full network in an end-to-end manner. While most computations usually come from conv layers, parameters easily explode when neurons are fully connected. It is important to select discriminative information in various deep structures.

In our experiments on general and domain specific datasets (ImageNet, CIFAR100, Adience and LFWA), we show how the proposed method leads to great complexity reductions. It brings down the total VGG16 size by 98\%-99\% and that of the compact InceptionNet (a.k.a. GoogLeNet) by up to 82\% without much accuracy loss (\textless 1\%). The corresponding FLOP reduction rates are as high as 83\% and 62\%, respectively. Additionally, we are able to derive more accurate models at lower complexities. Take age recognition on Adience for example, one model is over 3\% more accurate than the original net but only about 1/3 in size. Also, we compare the method with some of today's successful pruned/compact nets, such as MobileNet, SqueezeNet,~\cite{han20150},~\cite{li2016pruning},~\cite{molchanov2019} and show the value of deep discriminative pruning. Finally, in the above cases at least, we experimentally show that the fewer unrelated and interfering parameters the model has, the better it can generalize to unseen test data, and the less likely the model will be hit by adversarial attacks and noises (e.g. FGSM, Newton, Gaussian, Poisson, speckle).

\section{Related Work}

Early approaches to artificial neural networks pruning date back to the late 1980s. Some pioneering examples include magnitude-based biased weight decay~\citep{pratt1989}, Hessian based Optimal Brain Damage~\citep{lecun1989prune} and Optimal Brain Surgeon~\citep{hassibi1993}. Since those approaches are aimed at shallow nets, assumptions that were made, such as a diagonal Hessian in~\cite{lecun1989prune}, are not necessarily valid for deep neural networks. Please refer to~\cite{reed1993} for more early approaches.

In recent years, with increasing network depths comes more complexity, which reignited research into network pruning.
\cite{han20150} discard weights of small magnitudes. Small weights are set to zero and are masked out during re-training. Similarly, approaches that sparsify networks by setting weights to zero include~\cite{srinivas2015,mariet2016,jin2016,guo2016,hu2016,sze2017}.
With further compression techniques, this sparsity is desirable for storage and transferring purposes. That said, the actual model size and computation do not change much without specialized hardware and software optimization such as EIE~\citep{han2016eie}.
\cite{park2020} relate magnitude-based pruning to minimizing a single layer's Frobenius distortion induced by pruning. They develop lookahead pruning as a multi-layer generalization of magnitude-based pruning. In~\cite{frankle2019}, the authors hypothesize that within a large neural network (a bag of lottery tickets), there exists a small subnet (winning lottery ticket) that, when trained in isolation, can achieve similar accuracy. That said, the structure uncovered by pruning is experimentally shown to be harder to train from scratch.

Compared to pioneering pruning approaches based on individual weight magnitudes, filter or neuron level pruning has its advantages. Deep networks learn to construct hierarchical representations. Moving up the layers, high-level motifs that are more global, abstract, and disentangled can be built from simpler low-level patterns~\citep{bengio2013,zeiler2014}. In this process, the critical building block is the filter/neuron, which, through learning, is capable of capturing patterns at a certain scale of abstraction. Higher layers are agnostic as how the patterns are activated (w.r.t. weights, input, activation details).
Single weights-based approaches run the risk of destroying crucial patterns. For example, given uniform positive inputs, many small negative weights may jointly counteract large positive weights, resulting in a dormant neuron state. Single weight pruning based on magnitude would discard all small negative weights before reaching the large positive ones, reversing the neuron state. Inner-filter relationship is especially fragile at high pruning rates. 
Furthermore, instead of setting zeros in weights matrices, filter or neuron pruning removes rows, columns, depths in weight/convolution matrices, leading to direct space and computation savings on general hardware.

Early works in neuron/filter/channel pruning include~\cite{anwar2015,polyak2015,li2016pruning,tian2017,louizos2017,luo2017thinet,he2017channel}. They not only reduce the requirements of storage space and transportation bandwidth, but also bring down the initially large amount of computation in conv layers. Furthermore, with fewer intermediate feature maps generated and consumed, the number of slow and energy-intensive memory accesses is decreased, rendering the pruned nets more amenable to implementation on mobile devices. 
\cite{anwar2015} locate pruning candidates via particle filtering and introduce structured sparsity at different scales.
\cite{li2016pruning} equate filter utility to absolute weights sum. \cite{polyak2015} propose a `channel-level' acceleration algorithm based on unit variances. However, unwanted variances and noise may be preserved or even amplified. In~\cite{louizos2017}, the authors use hierarchical priors to prune nodes instead of single weights (and posterior uncertainties to determine fixed point precision). \cite{he2017channel} effectively prune networks through LASSO regression based channel selection and least square reconstruction. \cite{luo2017thinet} prune on the filter level guided by the next layer's statistics. 
In~\cite{zhuang2018}, the authors use additional classification and reconstruction losses on intermediate layers to help increase intermediate discriminative power and to select channels. They aggregate weight importance (gradients w.r.t weights) within a filter as filter importance. However, small gradients do not necessarily indicate low utility (e.g. at convergence). Similarly in~\cite{molchanov2019}, neuron importance is defined as the within-filter sum of weight importances (Taylor expansion of squared error induced without a weight). Although such methods prune on the filter level, there is still an implicit weight-level i.i.d assumption. In~\cite{molchanov2019}, it is an interesting idea that the pruning is done after several batches during retraining. That said, pruning importance based on only a few minibatches could be misleading and structures `greedily' pruned are unrecoverable.
\cite{lin2019} utilize generative adversarial learning to derive a pruning generator. During learning, they try to minimize the adversarial loss of a two-player game between the baseline and the pruned network, align the output of the two, and simultaneously encourage sparsity in the pruning soft mask.
In~\cite{he2019}, the authors point out two requirements of norm-based pruning, i.e. large norm deviation and small minimum norm. They then propose filter pruning via geometric median related redundancy (of filter norms in a layer) rather than importance.
Despite the promising pruning rates achieved by previous approaches, most of them possess one or both of the following drawbacks: (1) the utility measure for pruning, such as magnitudes or variances of weights or activation, is injected by human experts and is not directly related to task-dependent class separation. (2) utilities are often computed locally or considered on a local scale. Relationships within filter, layer, or across all layers may be missed.

In addition to pruning, there are some complementary and orthogonal approaches that can help constrain space and/or computational complexity. One is bit reduction such as weight quantization~\citep{rastegari2016} and Huffman encoding~\citep{han2015}. Some boost efficiency via decomposition of filters with a low-rank assumption, such as~\cite{denton2014exploiting,jaderberg2014speeding,zhang2016accelerating}. Another method is knowledge distillation~\citep{hinton2015distilling} where a small `student' model tries to achieve similar predicting power as a bigger `teacher' model on certain tasks. A bit of trial and error is usually involved in searching for the student net architecture. There are also some methods that utilize depth-wise separable convolution instead of the regular one to constrain model complexity~\citep{chollet2017xception,howard2017}. Last but not least, compact layers or modules with a random set of $1\times1$ filters are widely adopted to constrain dimensions, e.g. Inception nets~\citep{szegedy2015}, ResNets~\citep{he2016}, SqueezeNet~\citep{iandola2016}, MobileNet~\citep{howard2017}, and NiN~\citep{lin2014}. However, with an inappropriate filter number, it runs the risk of impeding information flow or increasing redundancy and interference. 
Most popular compact architectures are designed with human heuristics on some general datasets. Compared to fixed structures, a pruning method paying direct attention to the task utility in question can be flexible and can fit different task demands dynamically. This is desirable for a wide variety of real-world applications where adopting ImageNet suitable models have become a standard for industry best practices. As a matter of fact, in cases with limited data and strict timing requirements (e.g. car forward collision warning), a large, slow, and possibly overfitted model is hardly of any use. For visual classification, we define task utility as task dependent class separation power. In this paper, we capture it by deep LDA and use it to guide the pruning process. Our inspiration comes from neuroscience findings~\citep{mountcastle1957,valiant2006} which show that, despite the massive number of neurons in the cerebral cortex, each neuron typically receives inputs from a small task-dependent set of other neurons.
To our best knowledge, our method is the first one that employed LDA in deep dimension reduction. Unlike~\cite{dorfer2016}, no expensive optimization or extra transformation besides the network itself is needed.

\section{Task-dependent Deep Fisher LDA Pruning}\label{sec:ourmethod}

In this paper, we propose a neuron-level deep LDA pruning approach that pays direct attention to final task-dependent class separation utility and its holistic cross-layer dependency. We treat pruning as discriminative dimensionality reduction in the deep feature space by unravelling factors of variation and discarding those of little or even harmful/interfering utility.

The approach is summarized as Algorithm~\ref{alg:overall}.
As a pre-step, the base net is fully trained, with cross entropy loss, L2 regularization, and Dropout that helps punish co-adaptations. The main algorithm starts pruning by unravelling useful variances from the decision-making layer before tracing the utility backwards through deconvolution across all layers to weigh the usefulness of each neuron or filter. By abandoning less useful neurons/filters, our approach is capable of gradually deriving task-optimal structures with potential accuracy boosts.

\begin{algorithm}
    \SetAlgoLined
    \KwInput{base net, acceptable accuracy $t_{acc}$ or model complexity $t_{com}$}
    \KwResult{task-desirable pruned models}
    \vspace{0.025in}
    \KwPre{SGD optimization with cross entropy loss, L2-regularization, and Dropout.}
    \While{accuracy $\geq$ $t_{acc}$ (or complexity $\geq$ $t_{com}$)}{
      \vspace{0.05in}
      \textbf{Step 1} $\rightarrow $ \textbf{Pruning}\\ \vspace{0.025in}
      \begin{algorithmic}
      {
          \STATE 1. Task Utility Unravelling from Final Latent Space \newline \hspace{0.5in} (Section~\ref{sec:geig})
          \STATE 2. Cross-Layer Task Utility Tracing via Deconv \newline \hspace{0.5in} (Section~\ref{sec:utilitytracing})
          \STATE 3. Pruning as Utility Thresholding (Section~\ref{sec:thresholding})
      }
      \end{algorithmic}
      \vspace{0.025in}
      \textbf{Step 2} $\rightarrow $ \textbf{Re-training}\\
      \begin{algorithmic}
      {
          \STATE Similar to the pre-training step. Save model if needed.
      }
      \end{algorithmic}
    }
    \caption{Deep LDA Pruning of Neural Network}
    \label{alg:overall}
\end{algorithm}
It is worth mentioning that the number of iterations needed in Algorithm~\ref{alg:overall} is related to task difficulty. For simple datasets, only one or two iterations are enough to achieve a high pruning rate without much accuracy loss while more iterations are needed for challenging tasks. We will dive into the main pruning step in the following subsections, with one subsection for each sub-step in Algorithm~\ref{alg:overall}.

\subsection{Task Utility Unravelling from Final Latent Space}\label{sec:geig}

We try to capture the task utility from the final latent space of a well-trained base net for a number of reasons: (1) This is the only place where task-dependent distinguishing power can be accurately and directly measured. All previous information feed to this layer. (2) Data in this layer are more likely to be linearly separable (only one decision-making FC layer left, softmax is just a post-decision monotonic normalization). This is a key assumption of LDA but often overlooked by many previous approaches. (3) Pre-decision neuron activations representing different motifs are shown empirically to fire in a more decorrelated manner than earlier layers. We will see how this helps shortly.

For each image, an M-dimensional firing vector can be calculated in the final hidden layer, which is called a firing instance ($M=4096$ for VGG16, $M=1024$ for Inception, pooling is applied when necessary). By stacking all such instances from a set of images, the firing data matrix $X$ for that set is obtained (useless 0-variance/duplicate columns are pre-removed). Our aim here is to abandon dimensions of $X$ that possess low or even negative task utility. Inspired by~\cite{fisher1936,belhumeur1997,yang2002kernel,hua2007,bekios2011}, Fisher's LDA is adopted to quantify this utility. Our goal of pruning is achieved by maximizing class separation through finding an optimal transformation matrix $W$:
\begin{equation} \label{eq:interintraratio}
W_{opt} = \underset{W}{\arg\max} \frac{\mid W^T\Sigma_bW \mid}{\mid W^T\Sigma_wW \mid}
\end{equation}
where

\begin{equation} \label{eq:intraclassscatter}
\Sigma_w = \sum_{i} \tilde{X_{i}}^T\tilde{X_{i}}
\end{equation}

\begin{equation} \label{eq:interclassscatter}
\Sigma_b = \Sigma_a - \Sigma_w
\end{equation}

\begin{equation} \label{eq:allclassscatter}
\Sigma_a = \tilde{X}^T\tilde{X}
\end{equation}

\noindent with $X_i$ being the set of observations obtained in the last hidden layer for category $i$,
$W$ linearly projects the data $X$ from its original space to a new space spanned by $W$ columns. The tilde sign ( $\tilde{}$ ) denotes a centering operation; For data $X$: 
\begin{equation}
\tilde{X} = (I_n - n^{-1} 1_n1_n^T) X
\end{equation}
\noindent where $n$ is the number of observations in $X$, $1_n$ denotes an $n\times1$ matrix of ones. Finding $W_{opt}$ in Equation~\ref{eq:interintraratio} involves solving a generalized eigenvalue problem:

\begin{equation} \label{eq:geigenvaluep}
\Sigma_{b} \vec{e_j} = v_j \Sigma_{w} \vec{e_j}
\end{equation}

\noindent where ($\vec{e_j}$,$v_j$) represents a generalized eigenpair of the matrix pencil $(\Sigma_{b},\Sigma_{w})$ with $\vec{e_j}$ as a $W$ column. If we only consider active neurons with non-duplicate pattern motifs, we find that in the final hidden layer, most off-diagonal values in $\Sigma_w$ and $\Sigma_b$ are very small.
In other words, aside from noise and meaningless neurons, the firings of neurons representing different motifs in the pre-decision layer are highly decorrelated (disentanglement of latent space variances,~\cite{bengio2013,zeiler2014}). It corresponds to the intuition that, unlike common primitive features in lower layers, higher layers capture high-level abstractions of various aspects (e.g. car wheel, dog nose, flower petals). The chances of them firing simultaneously are relatively low. In fact, different filter `motifs' tend to be progressively more global and decorrelated when navigating from low to high layers. The decorrelation trend is caused by the fact that coincidences or agreements in high dimensions can hardly happen by chance. Thus, we assume that $\Sigma_w$ and $\Sigma_b$ tend to be diagonal in the top layer. Since inactive neurons are not considered here, Eq.~\ref{eq:geigenvaluep} becomes:
\begin{equation} \label{eq:geigenvaluepplus}
(\Sigma{w}^{-1}\Sigma_{b}) \vec{e_j} = v_j \vec{e_j}
\end{equation}

\noindent According to Eq.~\ref{eq:geigenvaluepplus}, $W$ columns ($\vec{e_j}$, where $j = 0, 1, 2 ... , M'-1$, $M' \leqslant M$) are the eigenvectors of $\Sigma{w}^{-1}\Sigma_{b}$ (diagonal), thus they are standard basis vectors (i.e. $W$ columns and $M'$ of the original neuron dimensions are aligned). $v_j$s are the corresponding eigenvalues with: 
\begin{equation} \label{eq:vj}
\begin{split}
v_j = diag(\Sigma{w}^{-1}\Sigma_{b})_j = \frac{\sigma^2_b(j)}{\sigma^2_w(j)}
\end{split}
\end{equation}

\noindent where $\sigma^2_w(j)$ and $\sigma^2_b(j)$ are within-class and between-class variances along the $j$th neuron dimension. In other words, the optimal $W$ columns that maximize the class separation (Eq.~\ref{eq:interintraratio}) are aligned with ($M'$, a subset of) the original neuron dimensions. It turns out that when pruning, we can directly discard neuron $j$s with small $v_j$ (little contribution to Eq.~\ref{eq:interintraratio}) without much information loss. 
Thresholding strategies, such as the Otsu method~\cite{otsu1979threshold}, can be used to select $M'$ most discriminative neuron dimensions. In this paper, we choose the hyperparameter $M'$ via sensitivity analysis on validation data. We will keep the minimum number of top neuron dimensions that can maintain comparable accuracies based on freezed top latent space features. Section~\ref{sec:topneuronnum} demonstrates this procedure by examples.

\subsection{Cross-Layer Task Utility Tracing}\label{sec:utilitytracing}
\begin{figure*}
\begin{center}
\includegraphics[height=0.25\textheight]{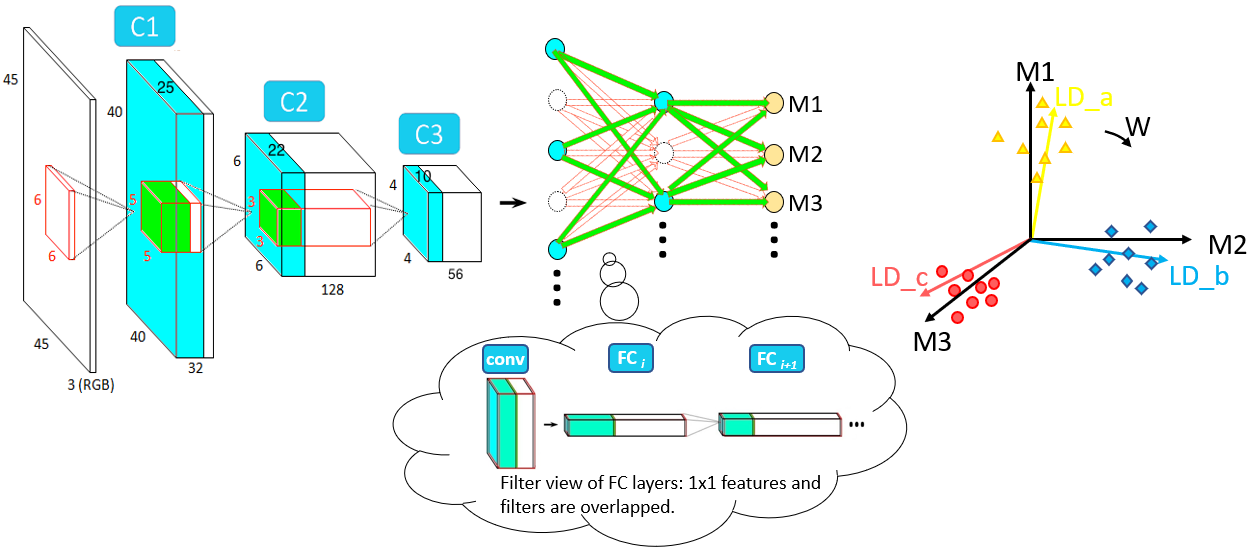}
\end{center}
\caption{Demo of neuron or filter level LDA-Deconv utility tracing. Useful (cyan) neuron outputs/features that contribute to final deep LDA utility through corresponding (green) next layer weights/filters, only depend on previous layers' (cyan) counterparts via deconv. White denotes useless components. $W$ is defined in Equation~\ref{eq:interintraratio}. The bubble cloud explains how deconv can be applied to FC layers. Each FC neuron is a stack of $1\times1$ filters with one $1\times1$ output feature map.}
\label{fig:prunedemo}
\end{figure*}
After unravelling twisted threads of deep variances and selecting dimensions of high task utility, the next step is to trace the class separation utility across all previous layers to guide pruning. Unlike local approaches, our pruning unit is concerned with a filter's/neuron's contribution to final class separation.

In signal processing, deconvolution (deconv) is used to reverse/undo an unknown filter's effect and recover corrupted sources~\citep{haykin1994blind}. Inspired by this, to recover each neuron/filter's utility, we trace the final discriminability, from the easily unravelled end, backwards across all layers via deconvolution.
In the final layer, only the most discriminative dimensions' response is preserved (other dimensions are set to 0) before deconv starts. 
In the ConvNet context, unlike convolution that involves sliding and dot products, deconvolution performs sliding and superimposing. There are many algorithms to compute or learn deconvolution. It is worth mentioning that `deconvolution' can be a confusing term today. Many deep learning frameworks define `deconvolution' as an up-scaling layer. It is a procedure where the weights are tuned for a particular purpose (e.g. segmentation). Unlike such frameworks, here we employee the `deconvolution' method as in~\citep{zeiler2014} and use the same terminology.
One difference is that Zeiler and Fergus~\citep{zeiler2014} use `deconvolution' for visualization purposes in the image space while we focus on reconstructing contributing sources over the layers.
Also, our method only back-propagates useful final variations. Irrelevant and interfering features of various kinds are `filtered out'.
As an inverse process of convolution, the unit deconv procedure is composed of unpooling (using max location switches), nonlinear rectification, and reversed convolution (using a transpose of the convolution Toeplitz-like matrix under an orthogonal assumption):
\begin{equation} \label{eq:deconv}
U_i = F_i^T Z_i
\end{equation}
Over the layers (ignoring nonlinearity and unpooling),
\begin{equation} \label{eq:crosslayerdeconv}
Z_{i-1}=U_i
\end{equation}
\noindent where $i$ indicates a layer, $U_i$ and $Z_i$ are layer $i$ input and output features with components not contributing to final utility removed. The $l$th columns of $U_i$ and $Z_i$ are respectively converted from layer $i$ reconstructed useful inputs and outputs w.r.t. input image $l$. 
Intuitively, Equation~\ref{eq:deconv} means performing convolution with the same filters transposed.
On the channel level:
\begin{equation} \label{eq:deconvchannelfeature}
U_{i,c} = \frac{1}{N}\sum_{l=1}^{N} \sum_{j=1}^{J_i} z_{i,j,l} * f_{i,j,c}^{t} 
\end{equation}

\noindent where `$*$' means convolution, $c$ indicates a channel, $N$ is the number of training images, $J$ is the feature map number, $f^t$ is a deconv filter that is determined after pre-training.
Based on Equation~\ref{eq:deconvchannelfeature}, we define deep LDA utility of layer $i$'s $c$th channel (or its producing filter in layer $i-1$) as:
\begin{equation} \label{eq:deconvfilter}
u_{i,c} = \max_{h,k}(U_{i,c}(h,k))
\end{equation}
\noindent where 
the maximum is taken over all spatial locations $(h,k)$. It means that as long as the corresponding filter spots something that finally contributes to classification separation, it is deserved to be kept no matter where the high utility occurs.
Our calculated dependency here is data-driven and is pooled over the training set, which models the established phenomenon in neuroscience which stipulates that multiple exposures are able to strengthen relevant connections (synapses) in the brain, i.e. the Hebbian theory~\citep{hebb2005}. It is worth mentioning that recovering or reconstructing the contributing sources to final class separation is not the same as computing a certain order parameter/filter dependency. Take 1st order gradient for example. Most parameters have 0 or small gradients at convergence, but it does not necessarily mean that these parameters are useless. Also, gradient dependency is usually calculated locally in a greedy manner. Structures pruned away based on a local dependency measure can never recover.

Fig.~\ref{fig:prunedemo} provides a high level view of cross-layer task utility tracing. To extend the deconv idea to FC layers, we consider FC layers as special conv structures where a layer's input and weights are considered as stacks of $1\times1$ conv feature maps and filters (completely overlapped as shown in the bubble cloud in Fig.~\ref{fig:prunedemo}). One difference is that FC structures do not have pooling layers in-between and no unpooling max switches are needed. Therefore, in a similar manner to normal conv layers, task utility can be successfully passed backwards across fully connected structures via deconv.
For modular structures, the idea is the same except that we need to trace dependencies, i.e. apply deconvolution, for different scales in a group-wise manner. Our full net pruning, (re)training, and testing are done end-to-end and are thus supported by most deep learning frameworks.

With all neurons'/filters' utility for final discriminability known, pruning simply becomes discarding structures that are less useful to final classification (e.g. structures colored white in Fig~\ref{fig:prunedemo}). Since feature maps (neuron outputs) correspond to next-layer filter depths (neuron weights), our pruning leads to filter-wise and channel-wise savings simultaneously.
In mathematical terms, input of conv layer $i$ can be defined as one data block $B_{data,i}$ of size $d_i\bigtimes m_{i} \bigtimes n_{i}$ meaning that the input is composed of $d_i$ feature maps of size $m_i \times n_i$ (from layer $i-1$). Parameters of conv layer $i$ can be considered as two blocks: the conv parameter block $B_{conv,i}$ of size $f_i \bigtimes c_i \bigtimes w_i \bigtimes h_i$ and the bias block $B_{bias,i}$ of size $f_i \bigtimes 1$, where $f_i$ is the 3D filter number of layer $i$, $c_i$ is the filter depth, $w_i$ and $h_i$ are the width and height of a 2D filter piece in that layer. It is worth noting that ${f}_{i-1}=d_i=c_i$. $B_{conv,i}(\sim,o,\sim,\sim)$ operates on $B_{data,i}(o,\sim,\sim)$, which is calculated using $B_{conv,i-1}(o,\sim,\sim,\sim)$ ($o$ is an ordinal number, `$\sim$' indicates all values along a dimension). When we prune away $B_{conv,i-1}(o,\sim,\sim,\sim)$, we effectively abandon the other two. In other words, $B_{conv,i}$ is pruned along the first and second dimensions simultaneously over the layers.

\subsection{Threshold Selection for Pruning}\label{sec:thresholding}
When pruning, layer $i-1$ neurons with a LDA-deconv utility score ($u_{i,c}$ in Eq.~\ref{eq:deconvfilter}) smaller than a threshold are deleted.
In an over-parameterized model, the number of `random', noisy, and irrelevant structures/sources explodes exponentially with depth. In contrast, well-trained cross-layer dependencies of final class separation are sparse. Unlike noise or random patterns, to construct a `meaningful' motif, we need to follow a specific path(s). It is this cross-layer sparsity of usefulness (task-difficulty-related) that greatly contributes to pruning, not just the top layer. To quickly get rid of massive less informative neurons while being cautious in high utility regions (at high percentiles), we set the threshold for layer $i$ as:

\begin{equation} \label{eq:threshold}
t_i = \eta \sqrt{\frac{1}{P_i-1} \sum_{j=1}^{P_i} (x_j - \overline{x_i})^2}
\end{equation}

\noindent where $\overline{x_i}$ is the average utility of layer $i$ activations, $x_j$ is the utility score of the $j$th activation, and $P_i$ is the total number of layer $i$ activations (space aware, those with 0 utility are ignored in Eq.~\ref{eq:threshold}). The assumption here is that the utility scores in a certain layer follow a Gaussian-like distribution.
The pruning time hyper-parameter $\eta$ is constant over all layers and is directly related to the pruning rate. We could set it either to squeeze the net as much as possible without obvious accuracy loss or to find the `most accurate' model, or to any possible pruning rates according to the resources available and accuracies expected. In other words, rather than a fixed compact model like SqueezeNet or MobileNet, we offer the flexibility to create models customized to different needs.
In our experiments, $\eta$ is linearly increased, say 0.1, 0.2, .... Human intervention is needed anytime Equation~\ref{eq:threshold} leads to too much or too little parameter reduction due to the imperfection of the Gaussian-like assumption.
Network capacity decreases with reduced filters and parameters. Generic fixed compact nets follow an ad-hoc direction by using random numbers and types of filters while our pruning selects filter dimensions according to current task demands and generates pruned models that are more invariant to task-unrelated factors. After pruning at each iteration, retraining with surviving parameters is needed.

\section{Experiments and Results}\label{sec:experiments}

\begin{figure}[!h]
\begin{center}
\includegraphics[width=0.825\linewidth,height=0.26\textheight]{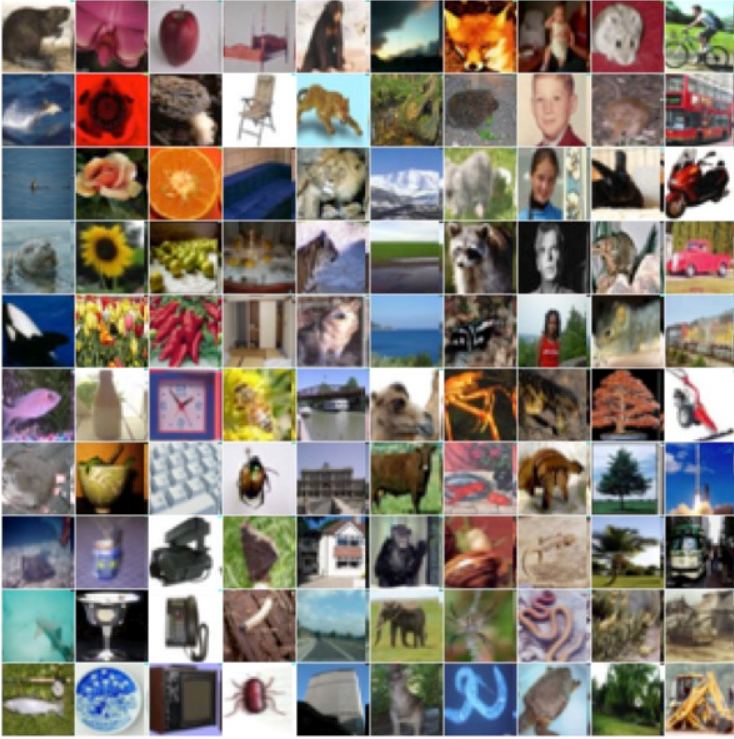}
\end{center}
\caption{Images from the CIFAR100 dataset representing different classes.}
\label{fig:cifar100examples}
\end{figure}

\begin{figure}[!h]
\centering 
\begin{subfigure}{0.115\linewidth}
    \includegraphics[width=\linewidth]{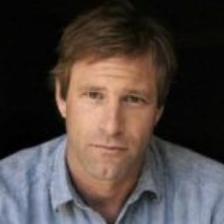}
\end{subfigure}
\begin{subfigure}{0.115\linewidth}
    \includegraphics[width=\linewidth]{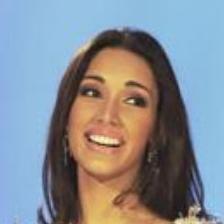}
\end{subfigure}
\begin{subfigure}{0.115\linewidth}
    \includegraphics[width=\linewidth]{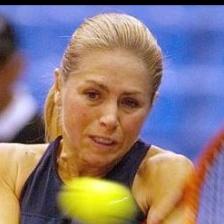}
\end{subfigure}
\begin{subfigure}{0.115\linewidth}
    \includegraphics[width=\linewidth]{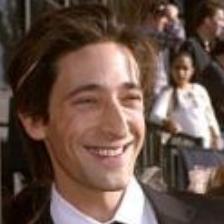}
\end{subfigure}
\begin{subfigure}{0.115\linewidth}
    \includegraphics[width=\linewidth]{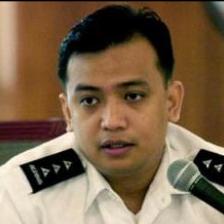}
\end{subfigure}
\begin{subfigure}{0.115\linewidth}
    \includegraphics[width=\linewidth]{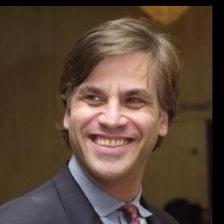}
\end{subfigure}
\begin{subfigure}{0.115\linewidth}
    \includegraphics[width=\linewidth]{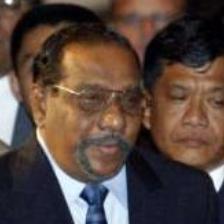}
\end{subfigure}
\begin{subfigure}{0.115\linewidth}
    \includegraphics[width=\linewidth]{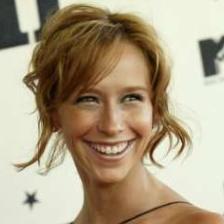}
\end{subfigure}
\caption{Images from the LFWA dataset (male/female, smiling/non-smiling examples).}
\label{fig:lfwexamples}
\end{figure}

\begin{figure}[!h]
\centering 
\begin{subfigure}{0.115\linewidth}
    \includegraphics[width=\linewidth]{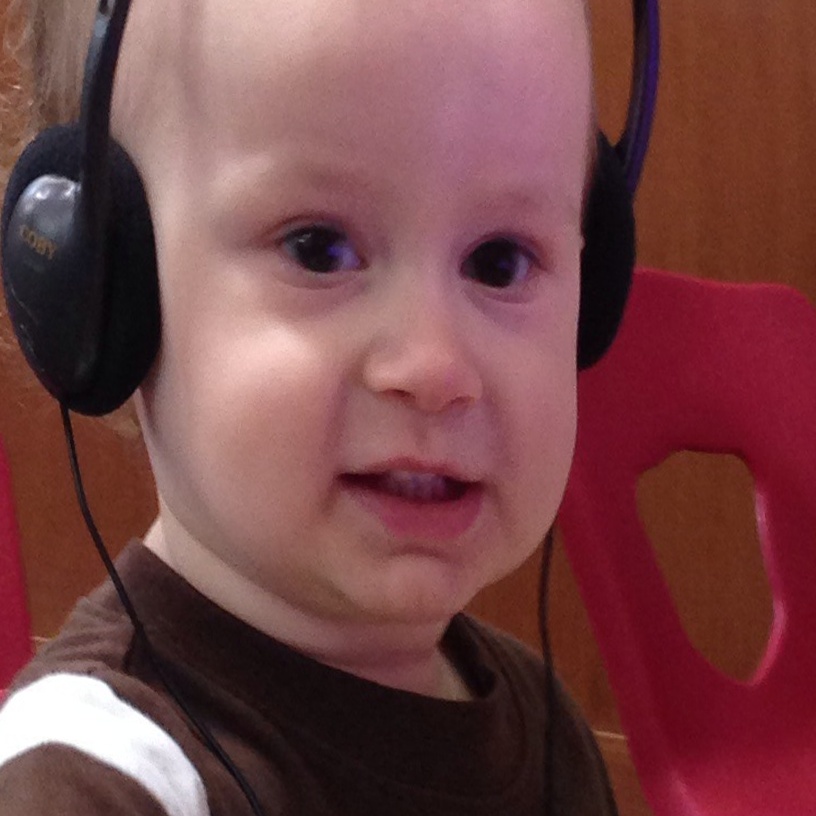}
\end{subfigure}
\begin{subfigure}{0.115\linewidth}
    \includegraphics[width=\linewidth]{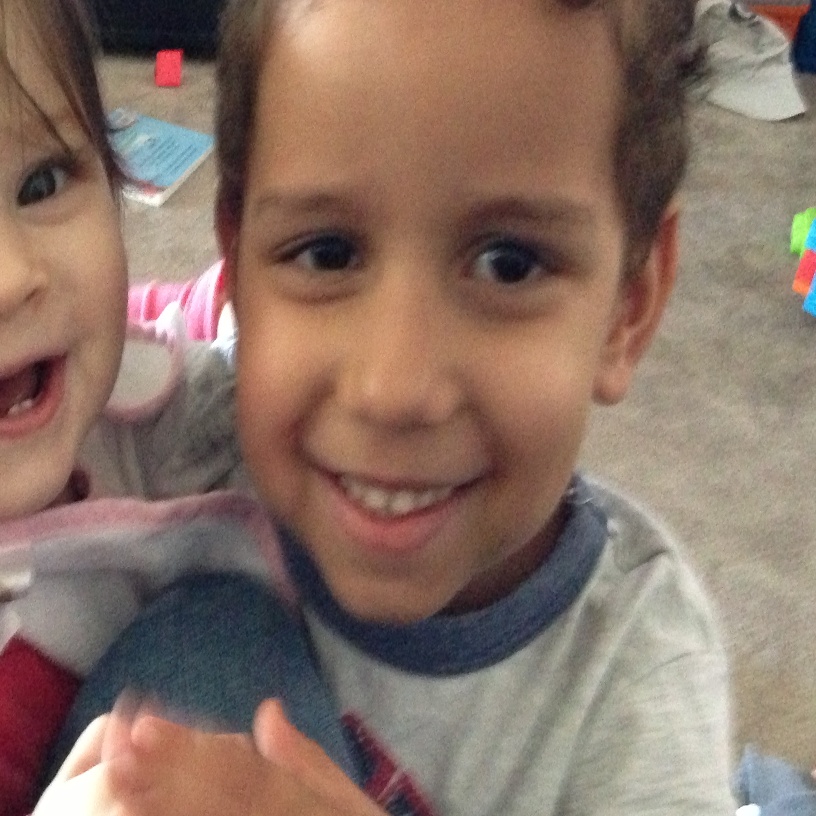}
\end{subfigure}
\begin{subfigure}{0.115\linewidth}
    \includegraphics[width=\linewidth]{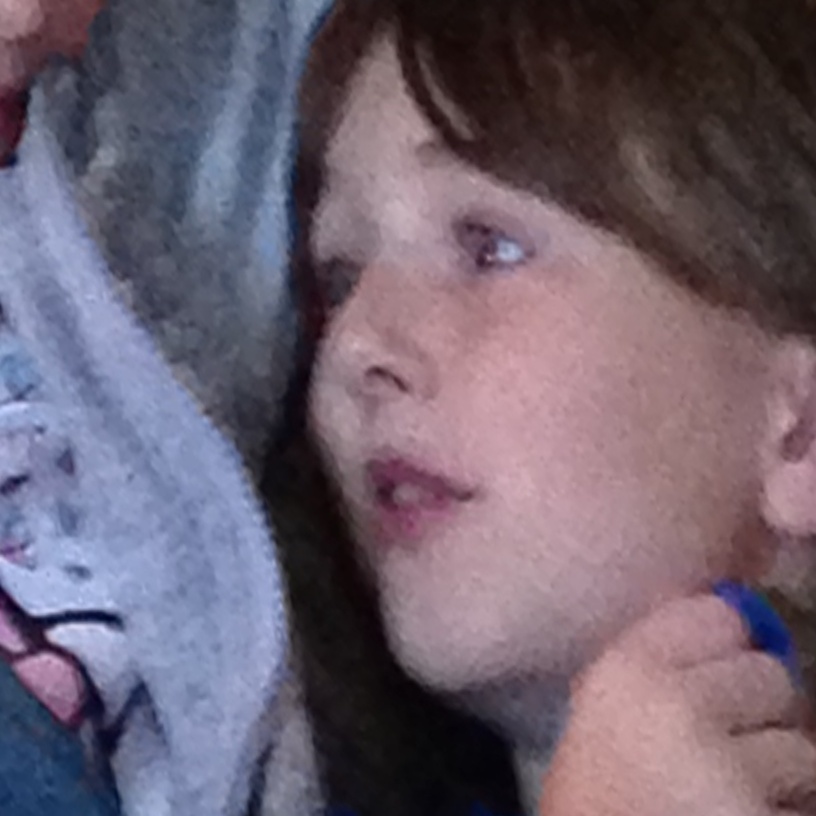}
\end{subfigure}
\begin{subfigure}{0.115\linewidth}
    \includegraphics[width=\linewidth]{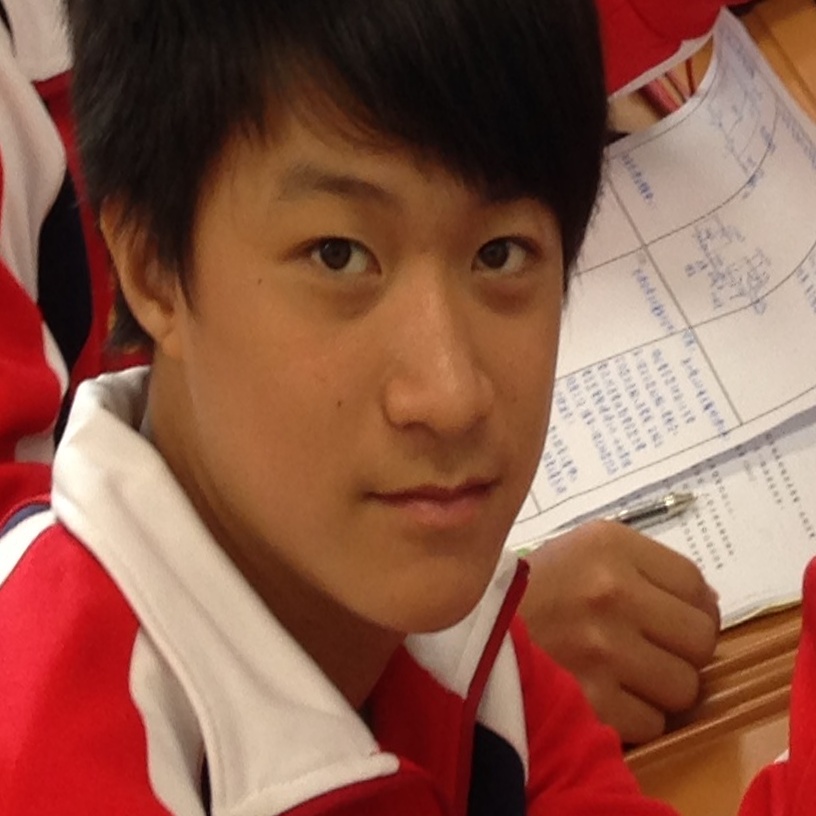}
\end{subfigure}
\begin{subfigure}{0.115\linewidth}
    \includegraphics[width=\linewidth]{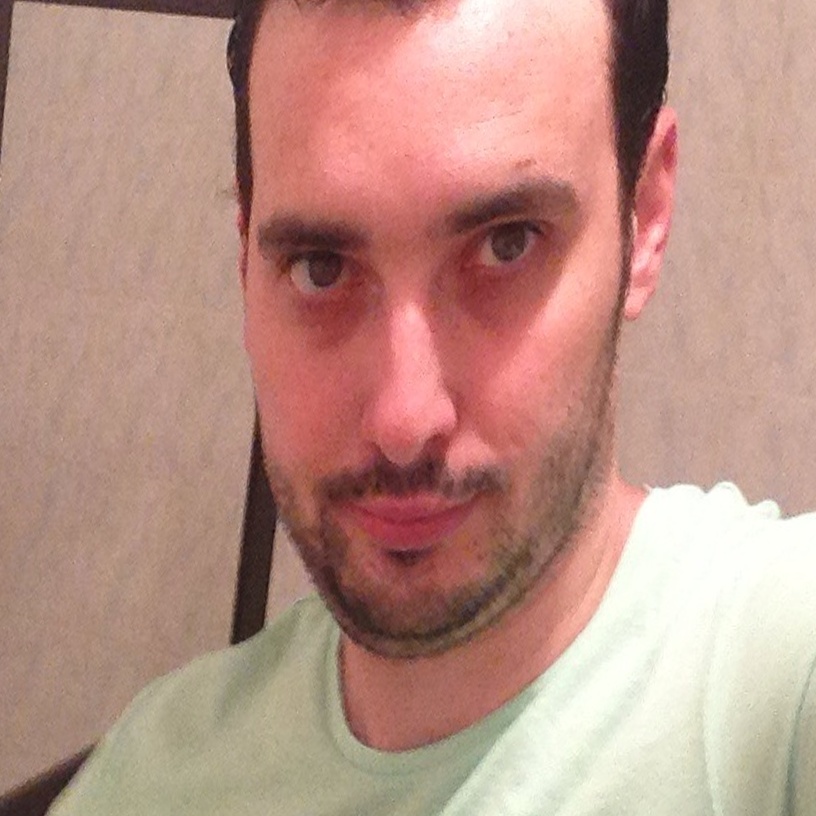}
\end{subfigure}
\begin{subfigure}{0.115\linewidth}
    \includegraphics[width=\linewidth]{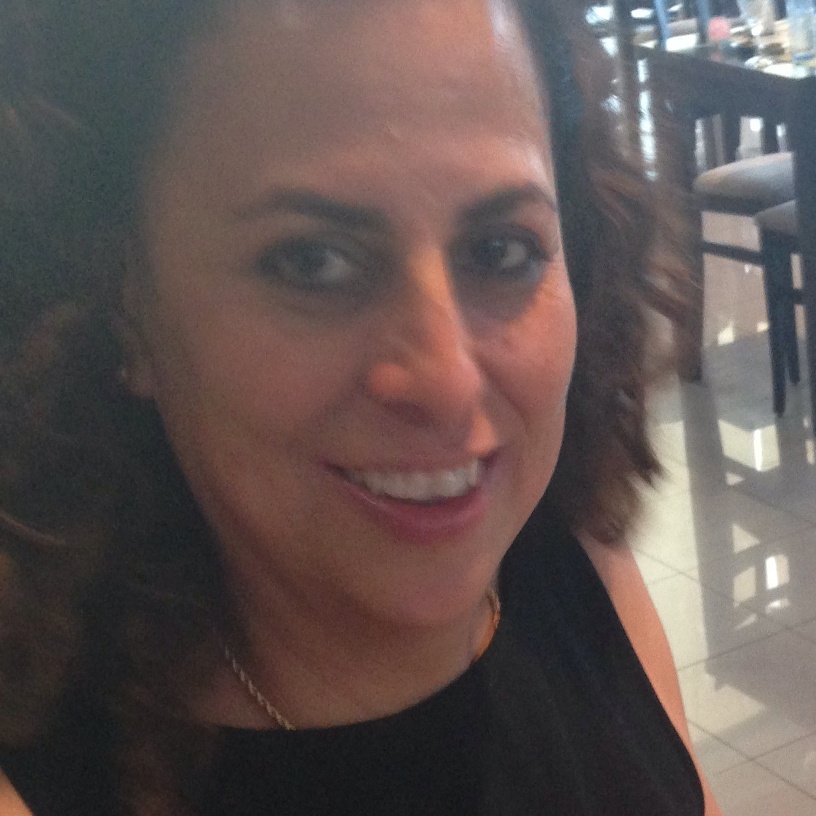}
\end{subfigure}
\begin{subfigure}{0.115\linewidth}
    \includegraphics[width=\linewidth]{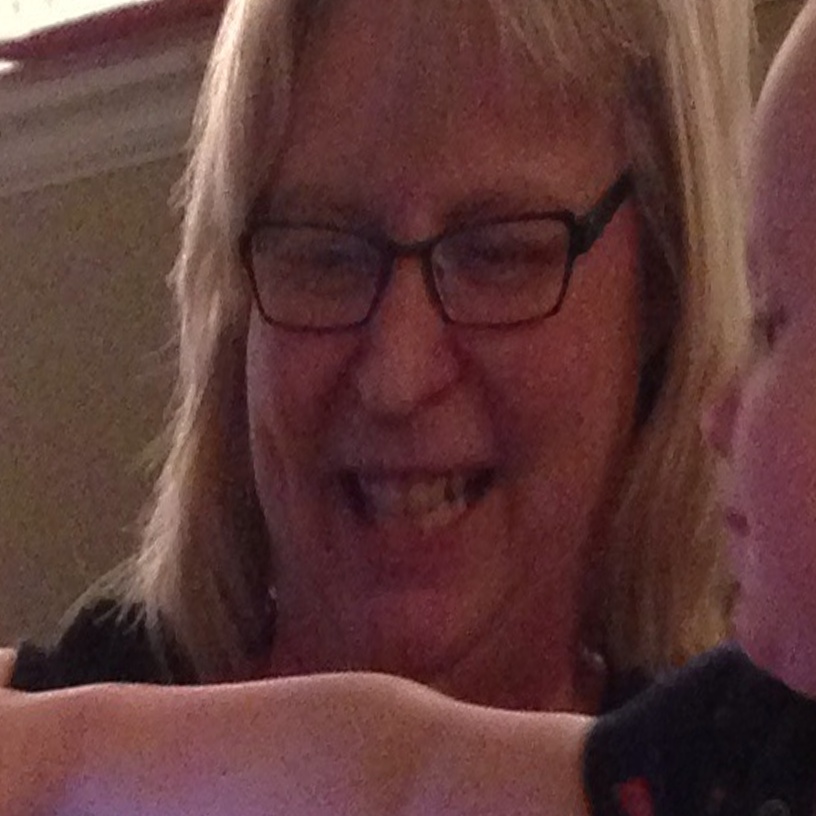}
\end{subfigure}
\begin{subfigure}{0.115\linewidth}
    \includegraphics[width=\linewidth]{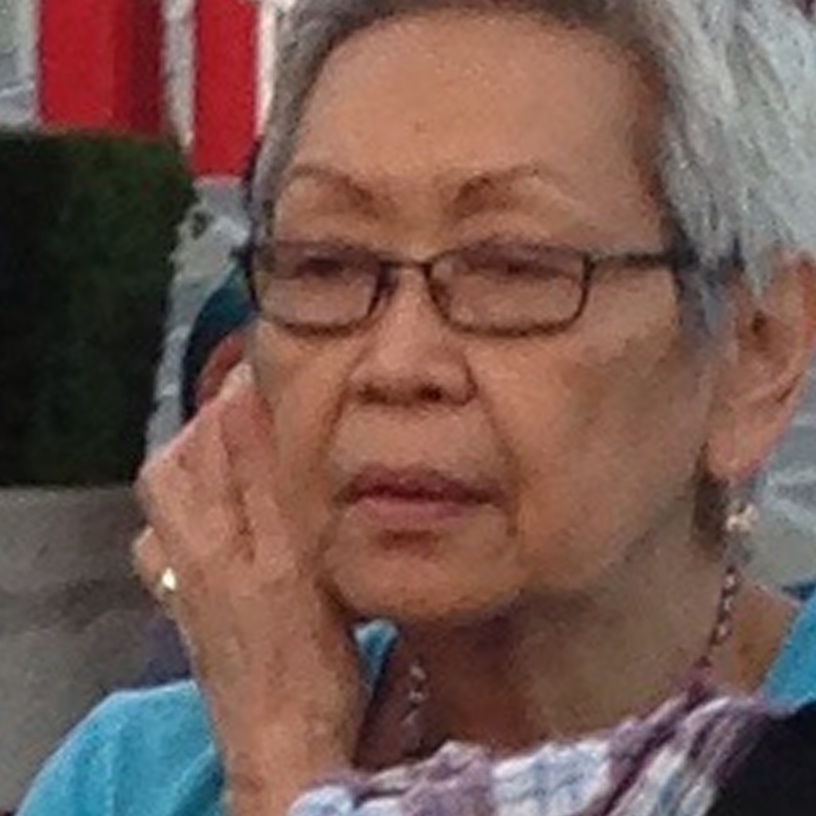}
\end{subfigure}
\caption{Images from the Adience Dataset representing different age groups.}
\label{fig:adienceexamples}
\end{figure}

In this paper, we use both conventional and module-based deep nets, e.g. VGG16~\citep{simonyan2015} and compact Inception net a.k.a GoogLeNet~\citep{szegedy2015}, to illustrate our deep LDA pruning method. Two general object datasets, i.e. ImageNet ILSVRC12~\citep{ILSVRC15} and CIFAR100~\citep{krizhevsky2009learning}, as well as two domain specific datasets, i.e. Adience~\citep{eidinger2014} and LFWA~\citep{liu2015} of facial traits, are chosen. 
Some most frequently explored attributes, such as age, gender, smile/no smile are selected from the latter two.
Non-ImageNet base models are pretrained on ILSVRC12 and are then fine-tuned on the new dataset. The suggested splits in~\cite{krizhevsky2009learning,levi2015,liu2015} are adopted. In addition, for CIFAR100, we use the last 20\% original training images in each of the 100 categories for validation purposes. For Adience, we use the first three folds for training, the 4th and 5th folds for validation and testing. For the LFWA dataset, we select identities with last name starting from `R' to `Z' for validation purposes. All images are pre-resized to the expected dimensions of the base net. Figure~\ref{fig:cifar100examples},~\ref{fig:lfwexamples} and~\ref{fig:adienceexamples} are some examples from the CIFAR100, LFWA, and Adience datasets. Please refer to~\url{http://www.image-net.org/} for example ImageNet images.


\subsection{Accuracy v.s. Pruning Rates}

This section demonstrates the relationship of accuracy change v.s. parameters pruned on the selected datasets. For comparison with our method, we include in the figures some other pruning approaches as well as modern compact structures, i.e. SqueezeNet~\citep{iandola2016} and MobileNet~\citep{howard2017}. We add the absolute base accuracy number to each figure just for reference. Many non-architecture factors can influence the absolute numbers (e.g. data augmentation, pre-processing, and optimization techniques). Our goal here is not the numbers themselves but rather their change due to pruning.

\subsubsection{LFWA}
\begin{figure}[!hb]
\centering 
\begin{subfigure}{0.48\linewidth}
    \includegraphics[width=\linewidth]{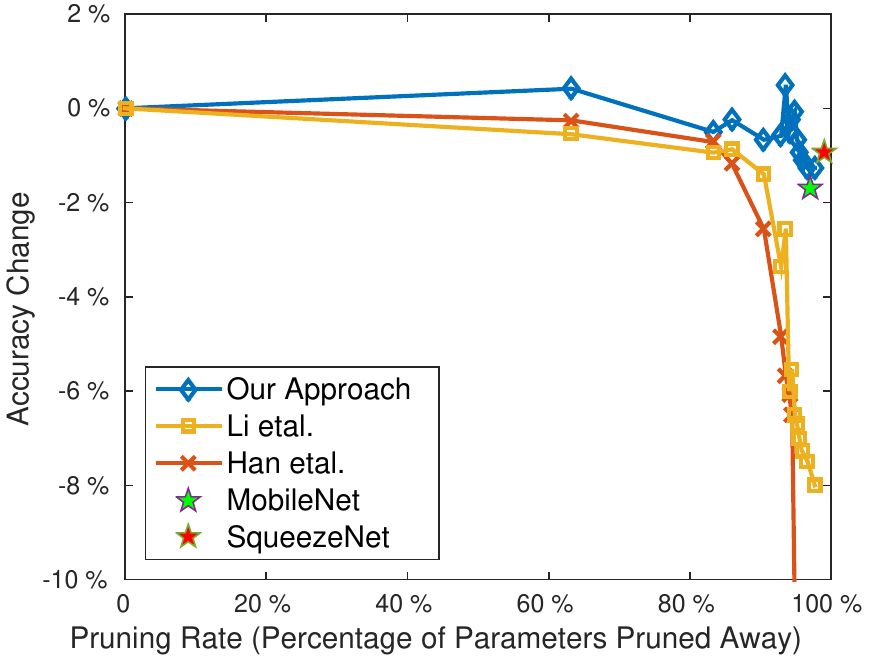}
    \caption{Gender, VGG16 base acc: 91.7\%}
     \label{fig:parammale}
\end{subfigure}
~ 
\begin{subfigure}{0.48\linewidth}
    \includegraphics[width=\linewidth]{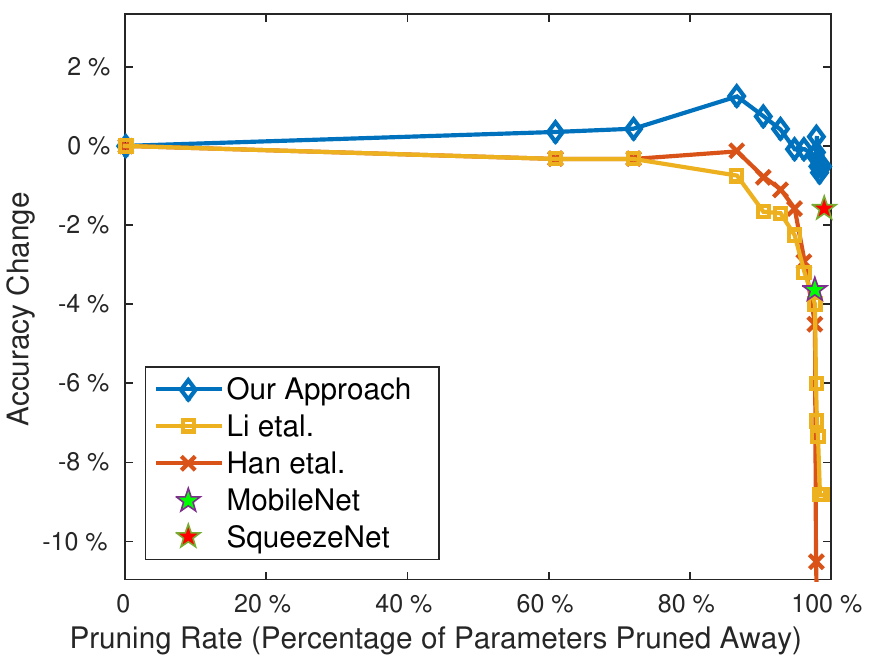}
    \caption{Smile, VGG16 base acc: 91.2\%}
    \label{fig:paramsmile}
\end{subfigure}
\caption{Accuracy change vs. parameters savings of our method (blue), \cite{han20150} (red), and \cite{li2016pruning} (orange) on LFWA validation data. For comparison, the performance of SqueezeNet~\citep{iandola2016} and MobileNet~\citep{howard2017} have been added. The `parameter pruning rate' for them implies the relative size w.r.t the original unpruned VGG16. In our implementation of~\cite{li2016pruning}, we adopt the same pruning rate as our method in each layer, rather than determine them empirically like the original paper does.}
\label{fig:vggparamcomparison}
\end{figure}
In LFWA, we choose gender and smile as example facial attributes since they are widely investigated and more interesting compared to others like color, shape, size of hair, nose, lip, beard, or the presence of glasses or jewellery.
One of the most popular conventional ConvNets VGG16 is adopted to test our approach's efficacy on the validation data (Fig.~\ref{fig:vggparamcomparison}). According to Fig.~\ref{fig:vggparamcomparison}, even with large pruning rates (98-99\%), our approach still maintains comparable accuracies to the original models (loss \textless 1\%). 
The other two methods~\cite{han20150} and~\cite{li2016pruning} suffer from earlier performance degradation, primarily due to their less accurate utility measures, i.e. single weights for~\cite{han20150} and sum of filter weights for~\cite{li2016pruning}. Additionally, for~\cite{han20150}, inner filter relationships are vulnerable to pruning especially when the pruning rate is large. This also explains why~\cite{li2016pruning} performs slightly better than \cite{han20150} at large pruning rates.

Moreover, higher accuracy is possible with less complexity. In the smile case for example, a 5x times smaller model can achieve 1.5\% more accuracy than the unpruned VGG16 net. Compared to the fixed compact nets, i.e. SqueezeNet and MobileNet, our pruning approach generally enjoys better performance at similar complexities. Even in the only pruning time exception in Fig.~\ref{fig:parammale} where Squeezenet has a slightly better accuracy than our pruned model of a similar size, much higher accuracies can be gained by simply adding back a few more parameters to our pruned net.

\begin{figure}[!h]
\begin{minipage}{\textwidth}
  \hspace{0.01in}
  \begin{minipage}{0.235\textwidth}
    \includegraphics[width=\textwidth,height=0.1275\textheight]{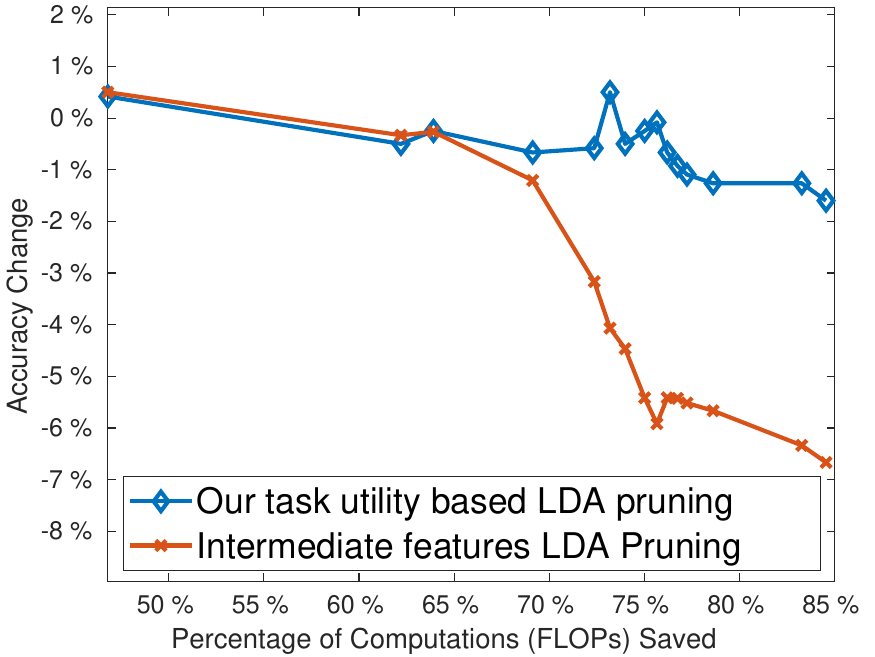}
  \end{minipage}
  \vspace{0.05in}
  \begin{minipage}{0.235\textwidth}
    \scriptsize
    \resizebox{\columnwidth}{!}{%
    \begin{tabular}{ccc}\hline
    FLOPs & Param\# & Acc Chg \\ \hline
    16B & 49M & +0.4\% \\
    11B & 19M & -0.2\% \\
    9.5B & 13M & -0.7\% \\
    8.2B & 8.6M & +0.5\% \\
    7.5B & 6.9M & -0.1\% \\
    7.4B & 6.5M & -0.7\% \\
    7.2B & 6.1M & -0.9\% \\
    5.2B & 3.1M & -1.0\% \\
    \hline
    \end{tabular}
    }
  \end{minipage}
  \\
  \begin{minipage}{0.235\textwidth}
    \includegraphics[width=\textwidth,height=0.1275\textheight]{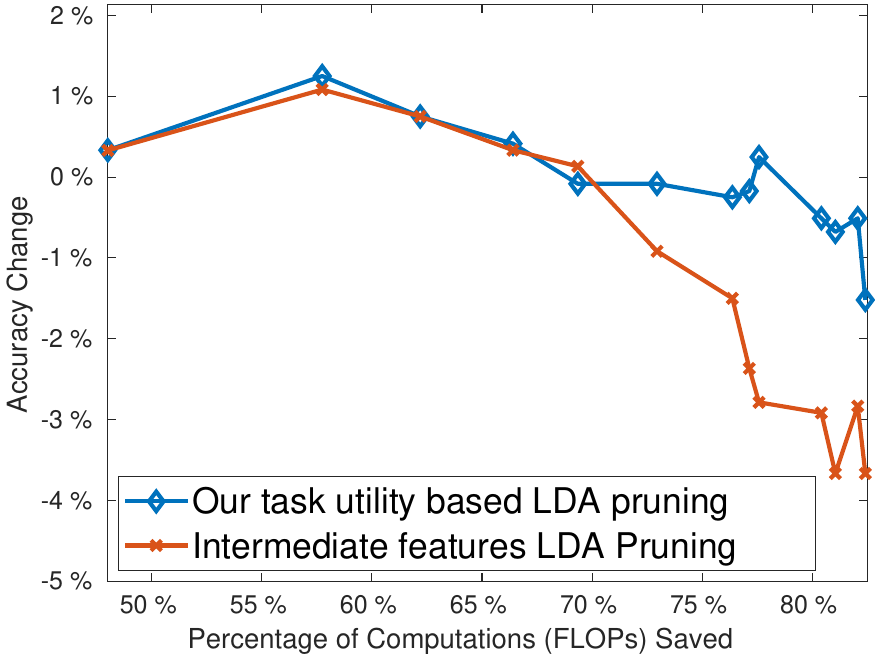}
  \end{minipage}
  \hspace{0.005in}
  \begin{minipage}{0.235\textwidth}
    \scriptsize
    \resizebox{\columnwidth}{!}{%
    \begin{tabular}{ccc}\hline
    FLOPs & Param\# & Acc Chg\\ \hline
    13B & 18M & +1.3\% \\
    12B & 13M & +0.8\% \\
    10B & 9.6M & +0.4\% \\
    8.3B & 5M & -0.1\% \\
    6.9B & 2.7M & +0.2\% \\
    6.0B & 2.5M & -0.5\% \\
    5.5B & 1.8M & -0.5\% \\
    5.4B & 1.7M & -1.5\% \\
    \hline
    \end{tabular}
    }    
  \end{minipage}
\end{minipage}
\hspace{0.01in}
\captionof{figure}{Accuracy change vs. FLOP savings of the method proposed in this paper (blue) and~\cite{tian2017} (red). Note: FLOPs are shared by both methods, Param\# and Acc Change are of the method in this paper. The top and bottom results are reported on LFWA gender and smile traits, respectively. Low pruning rates are skipped where the performance gap is small. The tables only show a few critical points in the corresponding curves on the left. Base model accuracies are the same as in Fig.~\ref{fig:vggparamcomparison}.}
\label{fig:flopcomparison}
\end{figure}

Also, we compare our approach with~\cite{tian2017} which applies linear discriminant analysis on intermediate conv features. The comparison (Fig.~\ref{fig:flopcomparison}) is in terms of accuracy vs. saved computation (FLOP) on the LFWA data. As in~\cite{han20150}, both multiplication and addition account for 1 FLOP. According to Fig.~\ref{fig:flopcomparison}, our method enjoys as high as 6\% more accuracy than~\cite{tian2017} at large pruning rates. The reasons are that our LDA pruning measure is computed where it directly captures final task classification power, the linear assumption is more easily met and the variances are more disentangled (so that direct neuron abandonment is justified, Section~\ref{sec:geig}).

\begin{table*}[!ht]
  \centering
  \captionsetup{width=0.8\textwidth}
  \caption{Testing accuracies on LFWA. `AF': accuracy first model, `PF': param\# first model. In the last row, Param\# and FLOPs are of our pruned models'. Our pruned models' Param\#s are shared by~\citep{li2016pruning,han20150} and our pruned models' FLOPs are shared by~\citep{li2016pruning}.~\cite{han20150} has the same FLOPs as the base. The base's name and its testing accuracy are in Row 1 parentheses. Original param\# and FLOPs for VGG16, MobileNet, and SqueezeNet are about 138M, 4.3M, 1.3M and 31B, 1.1B, 1.7B, respectively. M=$10^6$, B=$10^9$. Test set data are used here.}
  \resizebox{0.8\textwidth}{!}{%
  \begin{tabular}{c|c|c|c|c}
    \hline
     \multirow{2}{*}{Methods \& Acc} 
    & \multicolumn{2}{c|}{LFWA Gender (VGG, 91\%)} & \multicolumn{2}{c}{LFWA Smile (VGG, 91\%)} \\
    \hhline{~----}
    & AF & PF & AF & PF\\
    \hline
    MobileNet~\citep{howard2017} & \multicolumn{2}{c|}{89\%} & \multicolumn{2}{c}{87\%}\\
     \hhline{-----}
    SqueezeNet~\citep{iandola2016} & \multicolumn{2}{c|}{90\%} & \multicolumn{2}{c}{88\%}\\
     \hhline{-----}
    Han~\etal~\citep{han20150} & 89\% & 83\% & 91\% & 81\%\\
    Li~\etal~\citep{li2016pruning} & 88\% & 85\% & 91\% & 83\%\\
    Our approach & 93\% & 92\% & 93\% & 90\%\\
    \hhline{-----}
    (Param\#,FLOP) & (6.5M,7.4B) & (3.1M,5.2B) & (18M,13B) & (1.8M,5.5B)\\
    \hline
  \end{tabular}
  }
  \label{tab:highestaccvgg}
\end{table*}

To assess the generalization ability on unseen data, we report in Table~\ref{tab:highestaccvgg} the testing set performance of two of our pruned models for each task: one achieves the highest validation accuracy (`accuracy first' or AF model) and the other is the lightest model that maintains \textless 1\% validation accuracy loss (`parameter first' or PF model). The competing structures are also included. We try to make competing pruned models of similar complexities (last row). From Table~\ref{tab:highestaccvgg}, it is evident that our approach generalizes well to unseen data (highest accuracies over most cases). Apart from the overfitting-alleviating effect, one reason is that the proposed deep LDA pruning helps the over-parameterized model forget about task-irrelevant details and thus boosts its invariance to task-unrelated factors and changes in the unseen test data. The superiority is more obvious in the `parameter first' case. This agrees with previous validation results.

\subsubsection{Adience}
\begin{figure}[!h]
\centering
\begin{subfigure}{0.65\linewidth}
    \includegraphics[width=\linewidth]{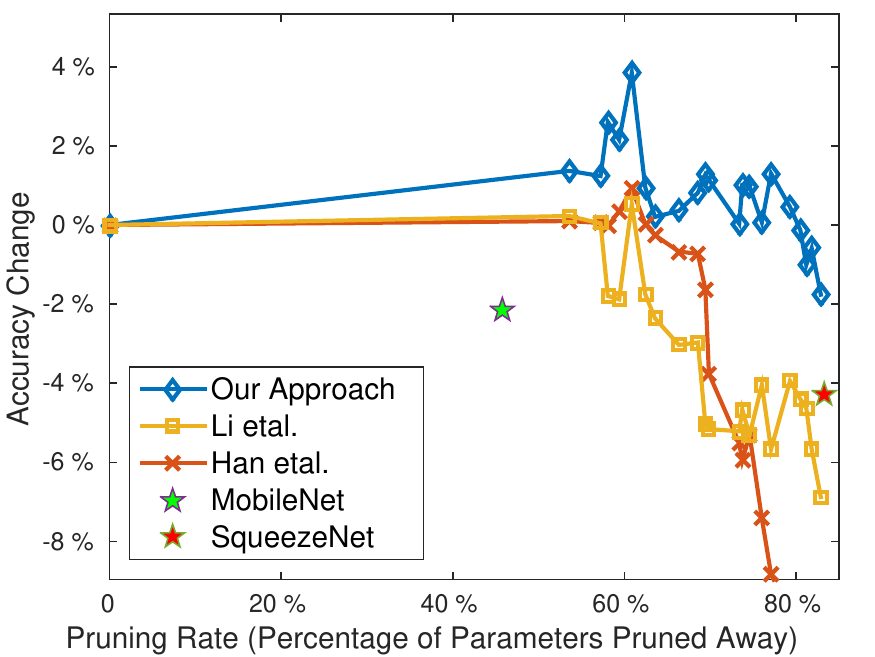}
    \caption*{Adience Age, Inception base accuracy: 48.9\%}
\end{subfigure}
\caption{Accuracy change vs. parameters savings of our method (blue), \cite{han20150} (red), and \cite{li2016pruning} (orange) on the Adience Age validation data. For comparison, the performance of SqueezeNet~\citep{iandola2016} and MobileNet~\citep{howard2017} have been added. The `parameter pruning rate' for them implies the relative size w.r.t the original unpruned Inception net. In our implementation of~\cite{li2016pruning}, we adopt the same pruning rate as our method in each layer, rather than determine them empirically like the original paper does.}
\label{fig:ageparamcomparison}
\end{figure}

In addition to the above-mentioned binary facial attributes, in this section, we show the accuracy vs pruning rate result on the multi-category age attribute from Adience. Inception net is employed as the base model.
We choose the Inception net over ResNets because the latter has human-injected dimension alignment. The skip/residual dimension has to agree with the main trunk dimension for summation. However, after pruning according to any importance measure (including ours), they do not necessarily agree without human intervention. Another reason is that, compared to residual models, inception nets offer us a wide range of filter types. By strategically selecting both the numbers and types of filters on different abstraction levels, we can derive task-desirable structures. 
Figure~\ref{fig:ageparamcomparison} shows the accuracy change vs pruning rate results of all competing methods on the validation split.
As we can see, comparable accuracy can be maintained even after throwing away over 80\% of the original Inception net parameters. During the pruning process, the proposed method obtains more accurate but lighter structures than the original net. For instance, a model of 1/3 the original size is 3.8\% more accurate than the original Inception net.
The gaps between our pruned models and fixed compact nets, i.e. MobileNet and SqueezeNet, are large because deep feature space dimension reduction with the goal to maximize final class separation is superior to reducing dimension using an arbitrary number of $1\times1$ filters. This supports the claim that pruning, or feature selection, should be task dependent. Also, the gaps between our pruned and fixed nets are wider compared to the VGG16 cases (Fig.~\ref{fig:vggparamcomparison}) for the reason that the method presented in this paper can take advantage of the filter variety in an inception module by strategically selecting both filter types and filter numbers according to task demands (more details in Sec.~\ref{sec:layerwisecomplexity}).
The testing set performance is reported in Table~\ref{tab:highestaccgoogleage}. The trends are similar as on the validation data.
\begin{table}[!h]
  \centering
  \captionsetup{width=0.5\textwidth}
  \caption{Testing accuracies on Adience Age. `AF': accuracy first model, `PF': param\# first model. In the last row, Param\# and FLOPs are of our pruned models'. Our pruned models' Param\#s are shared by~\citep{li2016pruning,han20150} and our pruned models' FLOPs are shared by~\citep{li2016pruning}.~\cite{han20150} has the same FLOPs as the base. The base's name and its testing accuracy are in Row 1 parentheses. Original param\# and FLOPs for InceptionNet, MobileNet, and SqueezeNet are about 6.0M, 4.3M, 1.3M and 3.2B, 1.1B, 1.7B, respectively. M=$10^6$, B=$10^9$. Test set data are used here.}
  \resizebox{0.5\textwidth}{!}{%
  \begin{tabular}{c|c|c}
    \hline
     \multirow{2}{*}{Methods \& Acc} 
    & \multicolumn{2}{c}{Adience Age (Inception, 55\%)} \\
    \hhline{~--}
    & AF & PF\\
    \hline
    MobileNet~\citep{howard2017} & \multicolumn{2}{c}{49\%}\\
     \hhline{---}
    SqueezeNet~\citep{iandola2016} & \multicolumn{2}{c}{50\%}\\
     \hhline{---}
    \cite{han20150} & 56\% & 43\%\\
    \cite{li2016pruning} & 56\% & 46\%\\
    Our approach & 58\% & 54\%\\
    \hhline{---}
    (Param\#,FLOP) & (2.3M,1.8B) & (1.1M,1.1B)\\
    \hline
  \end{tabular}
  }
  \label{tab:highestaccgoogleage}
\end{table}

\subsubsection{CIFAR100}

\begin{figure}[!h]
\centering 
\begin{subfigure}{0.65\linewidth}
    \includegraphics[width=\linewidth]{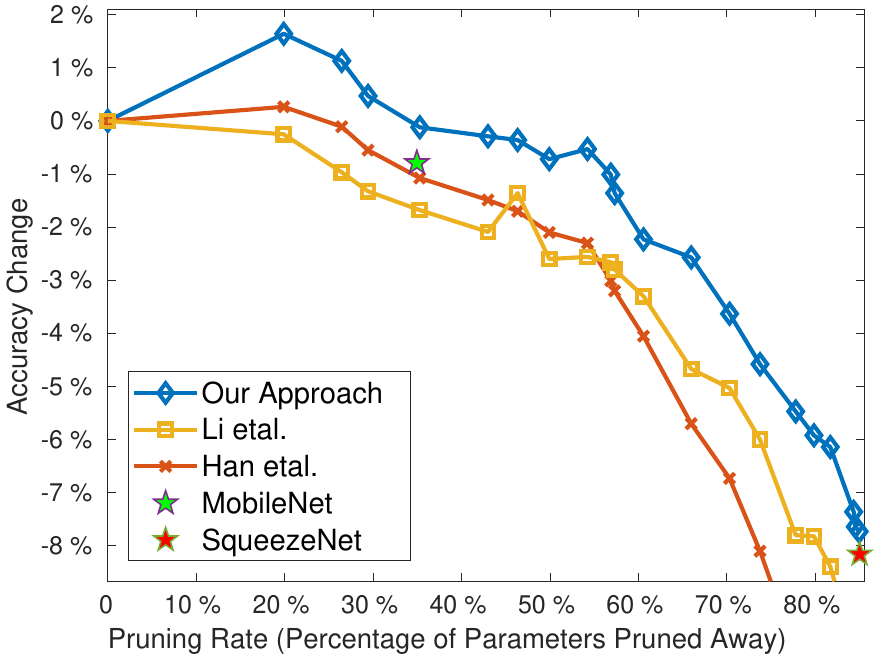}
    \caption*{CIFAR100, Inception base accuracy: 78.4\%}
    \label{fig:paramcifar100}
\end{subfigure}
\caption{Accuracy change vs. parameters savings of our method (blue), \cite{han20150} (red), and \cite{li2016pruning} (orange) on CIFAR100 validation data. For comparison, the performance of SqueezeNet~\citep{iandola2016} and MobileNet~\citep{howard2017} have been added. The `parameter pruning rate' for them implies the relative size w.r.t the original unpruned Inception net. In our implementation of~\cite{li2016pruning}, we adopt the same pruning rate as our method in each layer, rather than determine them empirically like the original paper does. Top-1 accuracy used.}
\label{fig:cifar100paramcomparison}
\end{figure}

The accuracy change against pruning rate on CIFAR100 is shown in Fig.~\ref{fig:cifar100paramcomparison}.
Top-1 accuracy is reported. Inception net is used as base.
As we can see, less than half of the total parameters (pruning rate 57\%) are able to maintain comparable accuracy and using about 80\% of the parameters leads to an accuracy that is nearly 2\% higher than the original. Additionally, although MobileNet and SqueezeNet perform similarly on Adience and LFWA, MobileNet performs clearly better on CIFAR100 mainly due to its suitable capacity in this particular case. This also indicates the superiority of providing a range of task-dependent models over fixed general ones. The former can help find the boundary between over-fitting and over-compression flexibly given a certain task.
Table~\ref{tab:highestaccgooglecifar100} shows the results on the test set.

\begin{table}[!ht]
  \centering
  \captionsetup{width=0.48\textwidth}
  \caption{Testing accuracies on CIFAR100. `AF': accuracy first model, `PF': param\# first model. In the last row, Param\# and FLOPs are of our pruned models'. Our pruned models' Param\#s are shared by~\citep{li2016pruning,han20150} and our pruned models' FLOPs are shared by~\citep{li2016pruning}.~\cite{han20150} has the same FLOPs as the base. The base's name and its testing accuracy are in Row 1 parentheses. Original param\# and FLOPs for InceptionNet, MobileNet, and SqueezeNet are about 6.1M, 4.3M, 1.3M and 3.2B, 1.1B, 1.7B, respectively. M=$10^6$, B=$10^9$. Test set data are used here.}
  \resizebox{0.48\textwidth}{!}{%
  \begin{tabular}{c|c|c}
    \hline
     \multirow{2}{*}{Methods \& Acc} 
    & \multicolumn{2}{c}{CIFAR100 (Inception, 78\%)}\\
    \hhline{~--}
    & AF & PF\\
    \hline
    MobileNet~\citep{howard2017} & \multicolumn{2}{c}{76\%}\\
     \hhline{---}
    SqueezeNet~\citep{iandola2016} & \multicolumn{2}{c}{71\%}\\
     \hhline{---}
    \cite{han20150} & 78\% & 73\%\\
    \cite{li2016pruning} & 78\% & 74\%\\
    Our approach & 80\% & 77\%\\
    \hhline{---}
    (Param\#,FLOP) & (4.8M,2.9B) & (2.6M,2.1B)\\
    \hline
  \end{tabular}
  }
  \label{tab:highestaccgooglecifar100}
\end{table}
\subsubsection{ImageNet}

For ImageNet, all images are resized to 256x256. During training, the images are randomly cropped to 224x224 and randomly mirrored about the vertical axis. Following the practice of most previous pruning works on ImageNet, we report accuracy change directly on the validation set (center crop is used). Here, we use a variant of Inception net that replaces 5$\times$5 conv layers with two 3$\times$3 conv layers. The first 3$\times$3 layer has the same filter number as its preceding 1$\times$1 conv layer and the second 3$\times$3 layer has the same number of filters as the original 5$\times$5 conv layer. This is the only architectural change we made. 
Later inception modules have more layers as well as some ad-hoc changes, such as larger input resolution (e.g. 299$\times$299), different filter distribution within modules and across layers, different configuration of stem layers. We chose not to incorporate those changes, in order to include as little human expert knowledge and handcrafting as possible. The objective would be to replace this type of architecture tweaking, many of which are not transferable to other tasks, with pruning.
Strictly speaking, the input to Inception V3 and V4 is not the same as the input of the competing fixed nets (e.g. MobileNet and SqueezeNet) since the $299 \times 299$ input contains more fine-grained information than the $224 \times 224$ input.

\begin{figure}[!h]
\centering 
\begin{subfigure}{0.65\linewidth}
    \includegraphics[width=\linewidth]{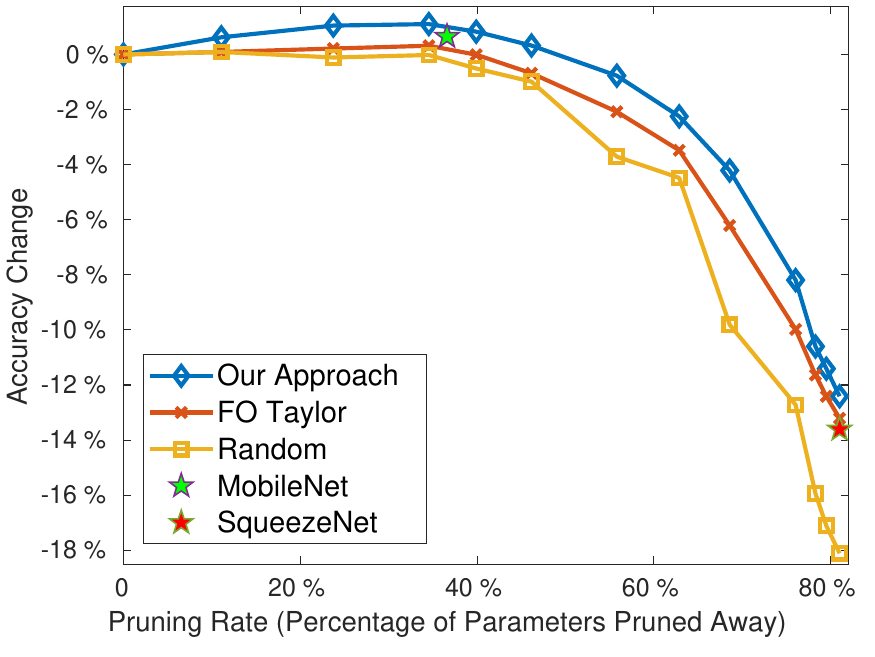}
    \vspace{-0.19in}
    \caption*{ImageNet, base model accuracy: 68.9\%}
\end{subfigure}
\caption{Accuracy change vs. parameters savings of our method (blue), FO Taylor~\cite{molchanov2019} (red), and random filter pruning (orange) on ImageNet. For comparison, the performance of SqueezeNet~\citep{iandola2016} and MobileNet~\citep{howard2017} have been added. The `parameter pruning rate' for them implies the relative size w.r.t the unpruned variant of Inception net (about 6.7M params). In our implementation of~\cite{molchanov2019} and random filter pruning, we adopt the same pruning rate as our method in each layer.}
\label{fig:imagenetparamcomparison}
\end{figure}

In this experiment on ImageNet, we compare our pruning with~\citep{molchanov2019} whose neuron importance measure is experimentally shown to be better than~\citep{han20150,li2016pruning}. We implement the FO Taylor measure of~\citep{molchanov2019} in Tensorflow as we do for the other methods and models (the original PyTorch implementation does not work beyond a certain pruning rate). We train the net to be pruned for one extra epoch, accumulate the importance scores over all training images, and prune after the end of the epoch. Results of random neuron/filter selection, SqueezeNet~\citep{iandola2016} and MobileNet~\citep{howard2017} are also reported. For the pruning methods, the same number of filters are selected by their corresponding neuron importance measure in a layer.
Figure~\ref{fig:imagenetparamcomparison} demonstrates the results.

As we can see from Fig.~\ref{fig:imagenetparamcomparison}, our pruning enjoys a high pruning rate even on the large ImageNet dataset and beats other competing approaches. A model with only 2.96M parameters (pruning rate $55.8\%$) and $2.0$ FLOPs is capable of maintaining accuracy comparable to the original. When the pruning rate is small, even the random filter selection can lead to satisfactory results. Generally speaking, the gap between our pruning method and the fixed nets (SqueezeNet and MobileNet) are small on ImageNet compared to the other datasets perhaps because the compact fixed nets are originally designed on ImageNet.
It is worth mentioning that we have also tested some tiny ResNets. All the resnets with fewer modules than ResNet10 (4 residual stages, each is a depth-2 conv block) have below-SqueezeNet accuracy, thus they are not reported in Fig.~\ref{fig:imagenetparamcomparison}.

\subsection{Layerwise Complexity Analysis}\label{sec:layerwisecomplexity}

\begin{figure*}[!h]
\centering
\begin{subfigure}{0.31\linewidth}
    \includegraphics[width=\linewidth,height=0.175\textheight]{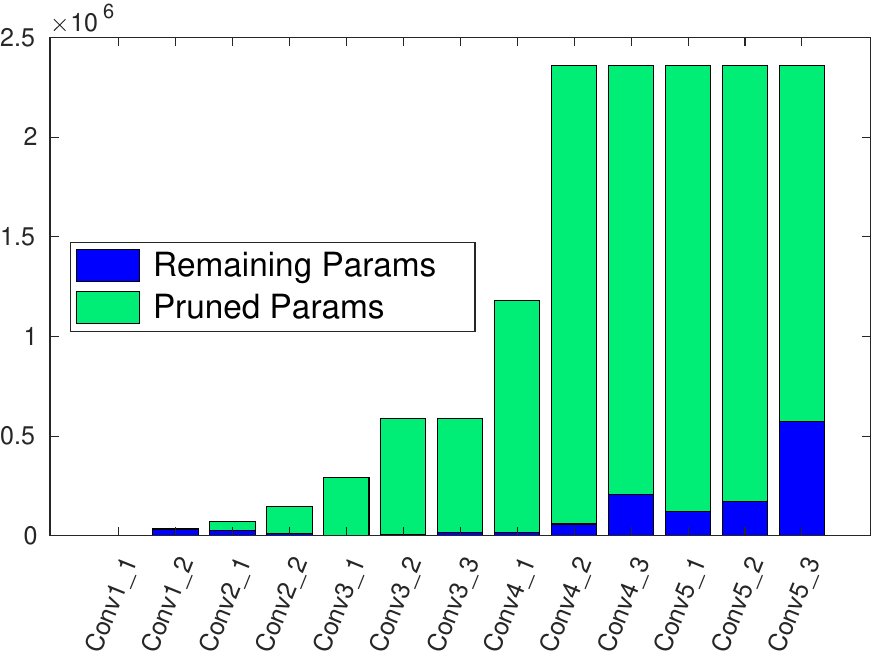}
    \caption{Conv layer param savings}
     \label{fig:parammalelayerwise}
\end{subfigure}
~ 
\begin{subfigure}{0.31\linewidth}
    \includegraphics[width=\linewidth,height=0.175\textheight]{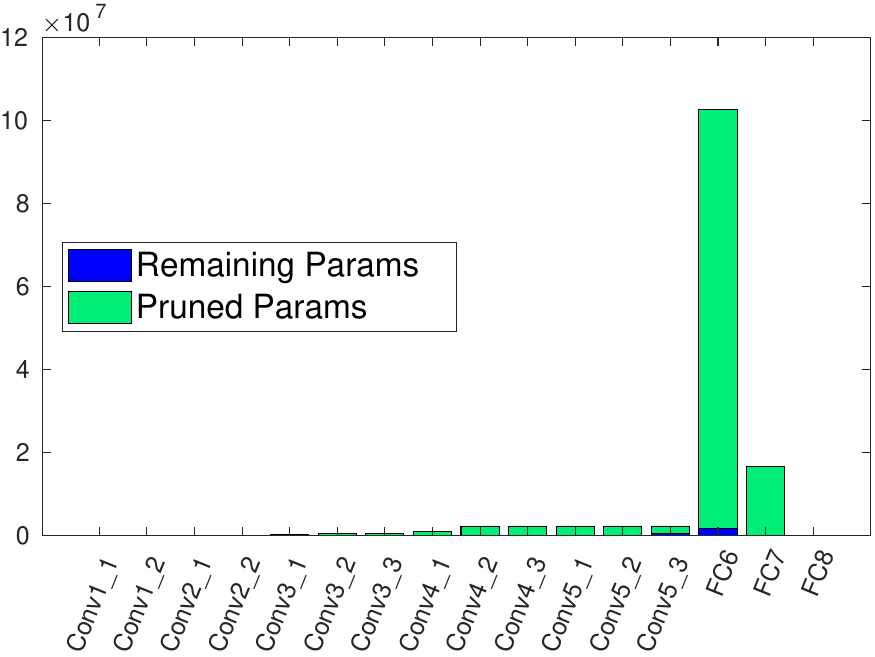}
    \caption{Param savings}
    \label{fig:parammalelayerwise_FC}
\end{subfigure}
~
\begin{subfigure}{0.31\linewidth}
    \includegraphics[width=\linewidth,height=0.175\textheight]{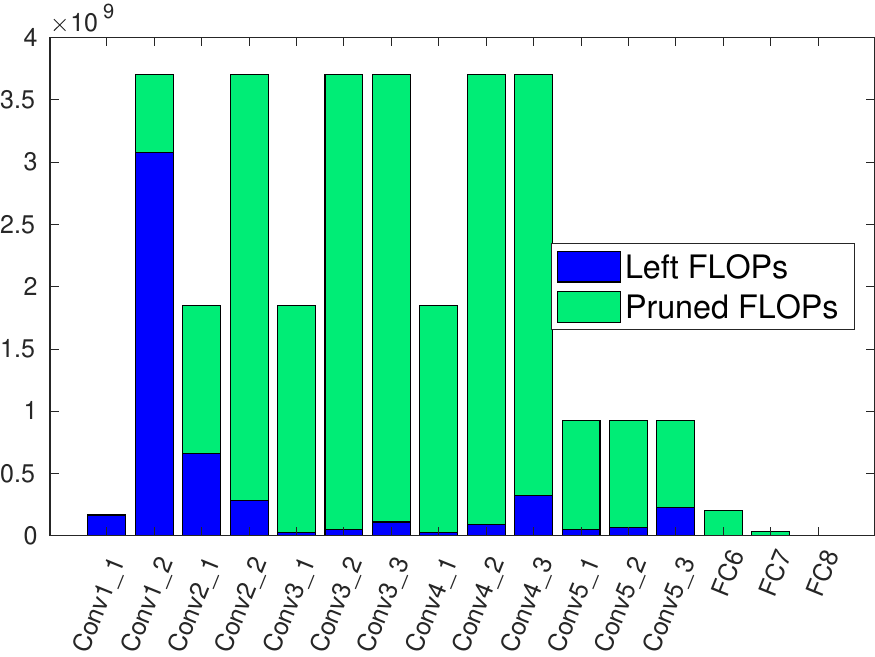}
    \caption{FLOP savings}
     \label{fig:flopmalelayerwise}
\end{subfigure}
\vspace{-0.1in}
\caption{Layerwise complexity reductions (LFWA gender, VGG16). Green: pruned, blue: remaining.}
\label{fig:layerwiseparamgender}
\vspace{0.15in}

\begin{subfigure}{0.31\linewidth}
    \includegraphics[width=\linewidth,height=0.175\textheight]{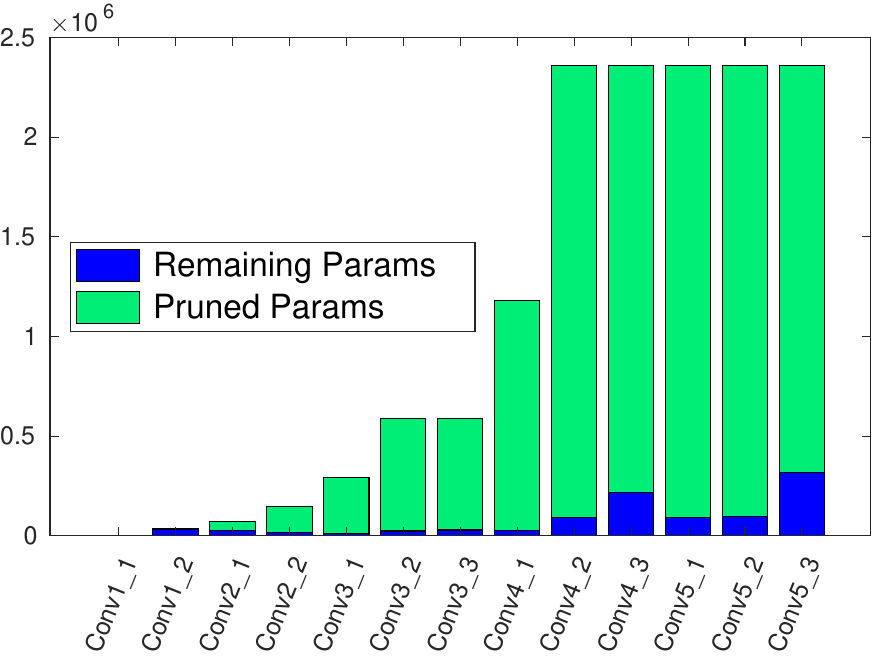}
    \caption{Conv layer param savings}
     \label{fig:paramsmilelayerwise}
\end{subfigure}
~ 
\begin{subfigure}{0.31\linewidth}
    \includegraphics[width=\linewidth,height=0.175\textheight]{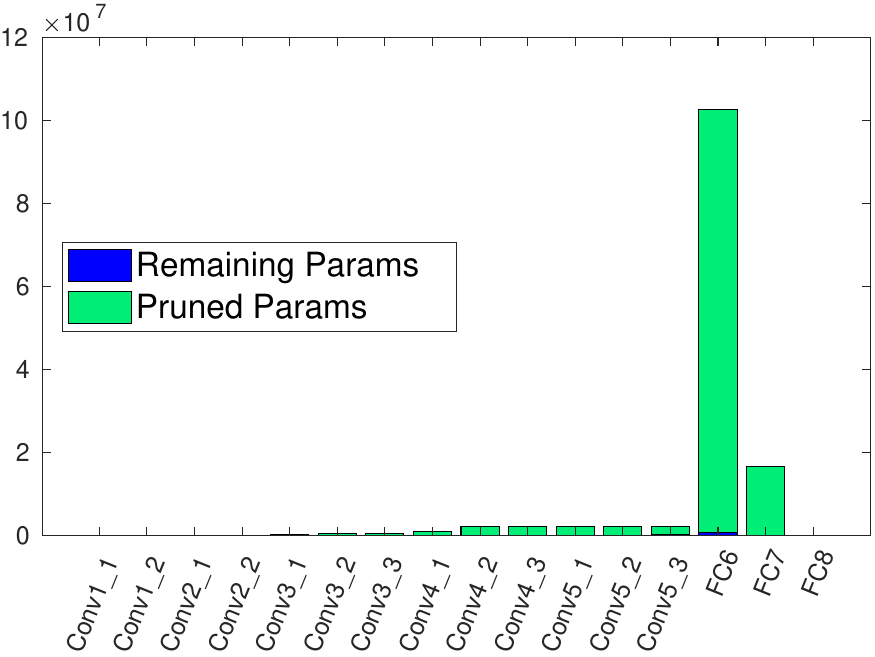}
    \caption{Param savings}
    \label{fig:paramsmilelayerwiseFC}
\end{subfigure}
~ 
\begin{subfigure}{0.31\linewidth}
    \includegraphics[width=\linewidth,height=0.175\textheight]{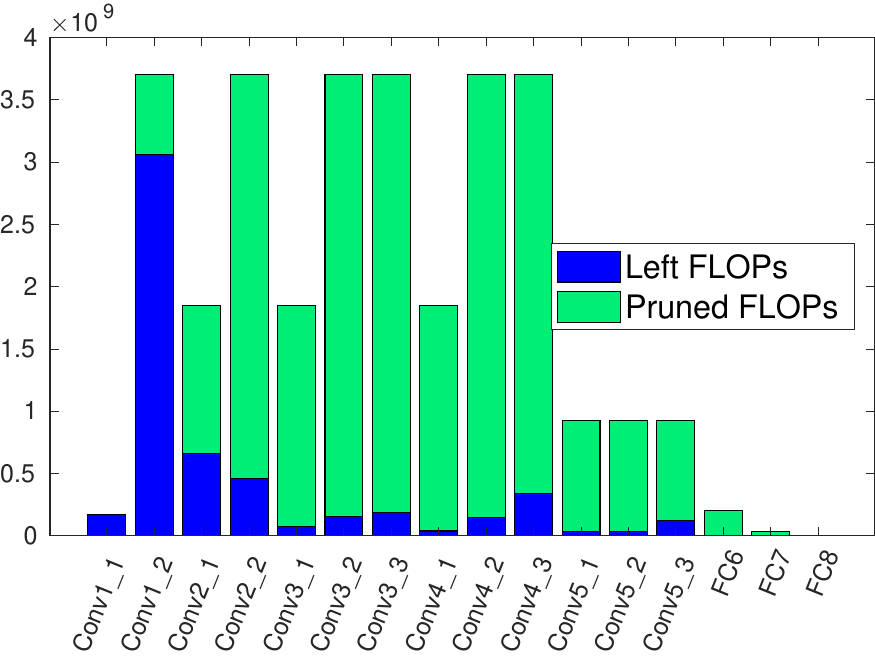}
    \caption{FLOP savings}
    \label{fig:flopsmilelayerwise}
\end{subfigure}
\vspace{-0.1in}
\caption{Layerwise complexity reductions (LFWA smile, VGG16). Green: pruned, blue: remaining.}
\label{fig:layerwiseparamsmile}
\vspace{0.15in}

\begin{subfigure}{0.45\linewidth}
    \includegraphics[width=\linewidth, height=0.2\textheight]{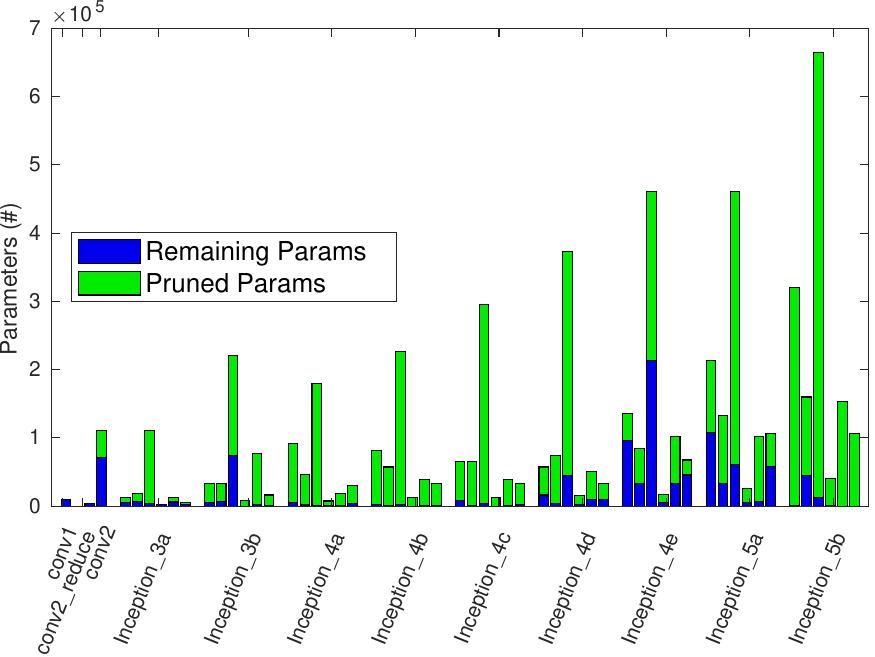}
    \caption{Param savings}
\end{subfigure}
~\hspace{0.25in}
\begin{subfigure}{0.45\linewidth}
    \includegraphics[width=\linewidth,height=0.2\textheight]{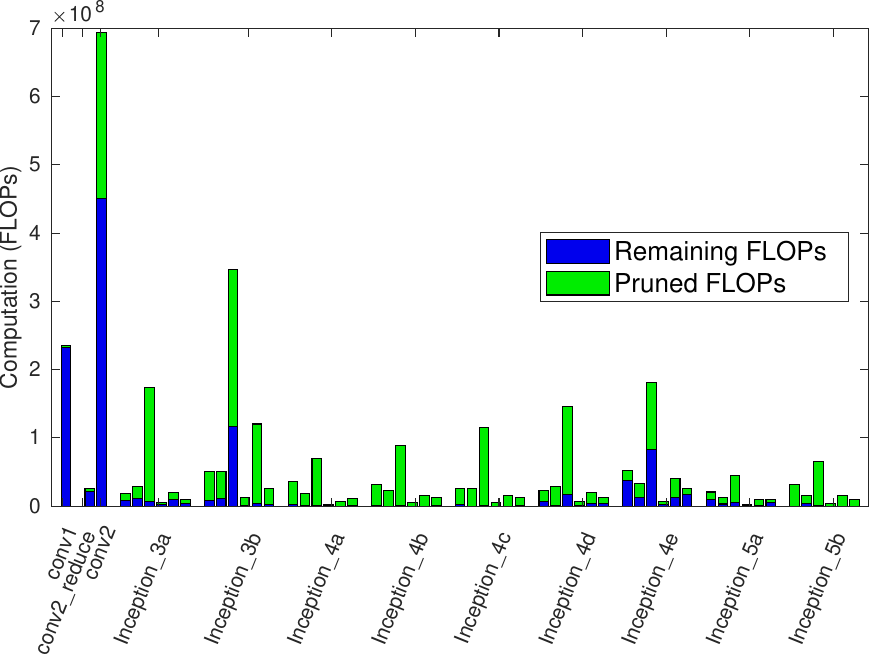}
    \caption{FLOP savings}
\end{subfigure}
\vspace{-0.1in}
\caption{Layerwise complexity reductions (Adience age, Inception). From left to right, the conv layers in a Inception module are ($1\times1$), ($1\times1$, $3\times3$), ($1\times1$, $5\times5$), ($1\times1$ after pooling). Green: pruned, blue: remaining.}
\label{fig:layerwiseparamage}
\vspace{0.15in}

\begin{subfigure}{0.45\linewidth}
    \includegraphics[width=\linewidth, height=0.2\textheight]{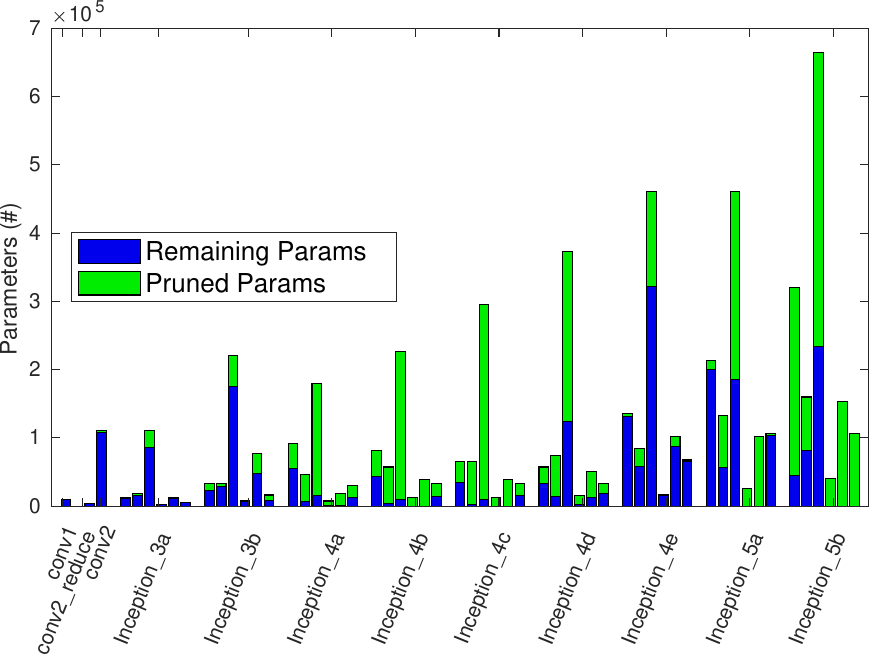}
    \caption{Param savings}
\end{subfigure}
~\hspace{0.25in}
\begin{subfigure}{0.45\linewidth}
    \includegraphics[width=\linewidth,height=0.2\textheight]{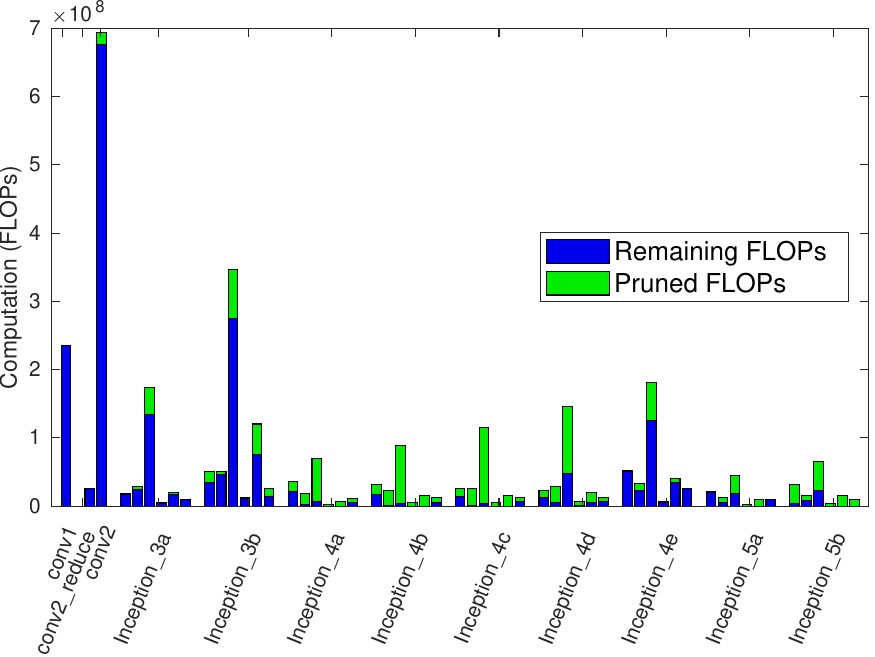}
    \caption{FLOP savings}
\end{subfigure}
\vspace{-0.1in}
\caption{Layerwise complexity reductions (CIFAR100, Inception). From left to right, the conv layers in a Inception module are ($1\times1$), ($1\times1$, $3\times3$), ($1\times1$, $5\times5$), ($1\times1$ after pooling). Green: pruned, blue: remaining.}
\label{fig:layerwiseparamcifar100}
\end{figure*}

\begin{figure*}[!h]
\centering
\begin{subfigure}{0.465\linewidth}
    \includegraphics[width=\linewidth, height=0.225\textheight]{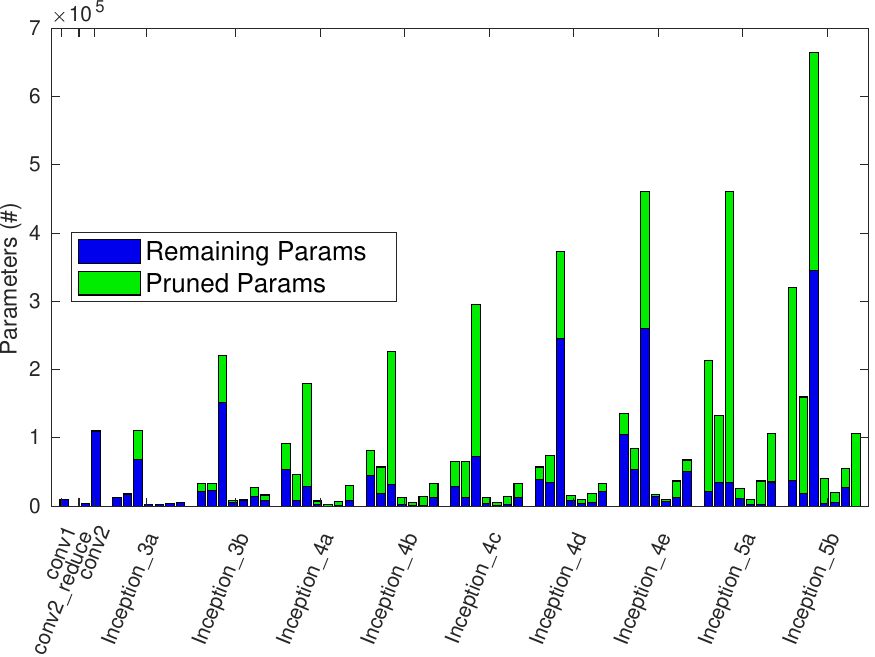}
    \caption{Param savings}
\end{subfigure}
~\hspace{0.25in}
\begin{subfigure}{0.465\linewidth}
    \includegraphics[width=\linewidth,height=0.225\textheight]{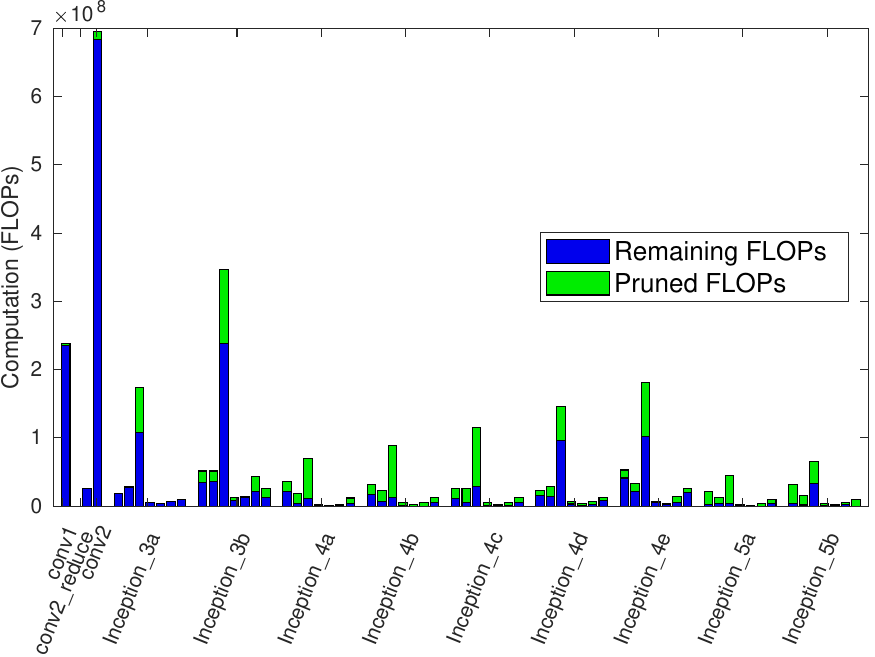}
    \caption{FLOP savings}
\end{subfigure}
\vspace{-0.1in}
\caption{Layerwise complexity reductions of the variant of Inception net on ImageNet. From left to right, the conv layers in a Inception module are ($1\times1$), ($1\times1$, $3\times3$), ($1\times1$, $3\times3$a, $3\times3$b), ($1\times1$ after pooling). Green: pruned, blue: remaining.}
\label{fig:layerwisecomplexityimagenet}
\vspace{0in}
\end{figure*}

Now we know that the proposed deep LDA pruning can find high performance deep models while being mindful of the resources available. In this section, we provide a more detailed layer-by-layer complexity analysis of our pruned nets in terms of parameters and computation. We consider fully-connected (dense) and conv layers. Fig.~\ref{fig:layerwiseparamgender},~\ref{fig:layerwiseparamsmile},~\ref{fig:layerwiseparamage},~\ref{fig:layerwiseparamcifar100},~\ref{fig:layerwisecomplexityimagenet} demonstrate layer-wise complexity reductions for the LFWA, Adience, CIFAR100, ImageNet cases respectively. The net we select for each case is the smallest one that preserves comparable accuracy to the original net.

Fig.~\ref{fig:layerwiseparamgender} and~\ref{fig:layerwiseparamsmile} show the LFWA cases with VGG16 as bases. Since the last conv layer output still has so many `pixels' that, when fully connected with the first FC layer's neurons, it generates a large number of parameters. With weight sharing, the number of conv layer parameters is limited. As a result, we add a separate parameter analysis for the conv layers. According to the results, our approach leads 
to significant parameter and FLOP reductions over the layers for the VGG16 cases. Specifically, the method effectively prunes away almost all the dominating FC parameters.

The base structure is the original InceptionNet for the Adience and CIFAR100 datasets and a slightly modified Inception net for ImageNet. As Fig.~\ref{fig:layerwiseparamage},~\ref{fig:layerwiseparamcifar100},~\ref{fig:layerwisecomplexityimagenet} show, a large proportion of parameters are pruned away. In each Inception module, different kinds of filters are pruned differently. This is determined by the scale where more task utility lies. By following a task-desirable direction, the method presented here attempts to maximize or maintain as much class separation power as possible when pruning. By choosing both the kinds of filters and the filter number for each kind, the approach also provides a feasible way to compact deep architecture design.

In the pruned models, most parameters in the middle layers have been discarded. In fact, the proposed method can collapse such layers to reduce network depth. In our experiments, when pruning reaches a threshold, all filters left in some middle modules are of size $1\times1$. They can be viewed as simple feature map selectors (by weight assignment) and thus can be combined and merged into the previous module's concatenation to form weighted summation. Such `skipping' modules pass feature representations to higher layers without incrementing the features' abstraction level. 
InceptionNet is chosen as an example because it offers more filter type choices without human-injected constraints on dimension alignment.
However, the proposed approach can be used to prune other modular structures as well, such as ResNets where the final summation in a unit module can be modeled as a concatenation followed by convolution.

In all layerwise-complexity figures above, the first few layers are not pruned very much. This is because earlier layers correspond to primitive patterns (e.g. edges, corners, and color blobs) that are commonly useful. In addition, early layers help sift out and provide some robustness to massive noisy statistics in the pixel space. Despite its data dependency, the proposed approach does not depend much on training `pixels', but pays more attention to deep abstract manifolds learned and generalized from training instances.
Overall, the pruned models are very light. On a machine with 32-bit parameters the models are respectively 11.9MiB, 6.7MiB, 4.1MiB, 10MiB, and 11.3MiB. During inference, they can fit into computer and cellphone memories or possibly even caches (with super-linear efficiency boosts).
In our experiments on a single Intel Xeon-E5 CPU core, one forward pass of the original model for each case takes 2.72s, 2.72s, 0.37s, 0.39s, and 0.29s, respectively. After the pruning, these numbers become 0.42s, 0.40s, 0.14s, 0.24s, 0.16s. Please note that the numbers are highly dependent on hardware specifics and are just for reference here.

\subsection{Model Robustness against Noises and Adversarial Attacks}

From the above sections, we can see that our pruning leads to great complexity reductions with a possibility of increasing prediction accuracy. In addition to efficiency boost and possible accuracy gain, in this section, we will investigate our pruning's effects on the model's robustness to input perturbations. 

One fair way to compare between the original and pruned models is to attack/modify the input data in the same way for both models and see how they react to the perturbation.
To this end, we first apply Gaussian, Poisson, speckle noises and two adversarial attacks, i.e. FGSM~\citep{goodfellow2014} and Newton Fool Attack~\citep{jang2017}, to the LFWA, Adience, CIFAR100 testing data and compare how the original and pruned models perform in terms of accuracy drops. Here, accuracy drop means accuracy difference between predicting on clean testing data and on noisy or attacked testing data using a model. In our experiment, the pruned model selected in each case has similar accuracy to the unpruned one on the clean test set. 
In the adversarial attack cases, the examples are generated from a third ResNet50 model (source model) and are transferred here as blackbox attacks to fool our models in comparison (target models). It is worth mentioning that we did not use any residual base structures for our pruning experiments.
In practice, not all attackers have the chance to make a large number of queries to the model being attacked or have access to the model details (e.g. gradients). So robustness analysis against transfer-based attacks is meaningful.
As mentioned in~\cite{bhagoji2018practical,huang2019black}, such transfer-based blackbox attacks are very common in the literature. Examples include~\citep{szegedy2014intriguing,goodfellow2015,papernot2016transferability,papernot2016practical,liu2016delving,carlini2017towards,moosavi2017universal,tramer2017space,madry2018towards,dong2018boosting,dong2019evading}.
The accuracy drop results of the original and pruned models due to the input perturbations are reported in Table~\ref{tab:noisegoogle} and~\ref{tab:noisevgg}, for Inception and VGG16 cases respectively.

As can be seen from the results, the pruned models are more, or at least equally, robust to the noises than corresponding original unpruned models. One possible reason is that with fewer task-unrelated random filters, the pruned models are less likely to pick up irrelevant noises and are thus less vulnerable. Also, as mentioned earlier, reducing parameters per se alleviates overfitting and thus brings down variance to data fluctuations. The deep nets are more prone to Gaussian and speckle noises than to Poisson noises.
Furthermore, we can see that our pruning method can help with model robustness to transfer-based adversarial attacks in many cases that we have investigated. This is because fewer irrelevant deep feature dimensions can possibly mean fewer breaches where the adversarial attacks can easily put near-boundary samples to the other side of the decision boundary. That said, the pruning's effect on robustness is less obvious in the simple FGSM cases as compared to the Newton Fool Attack cases. Overall, both the task and the net architecture have an influence on robustness. VGG16 and its pruned models are less susceptible to the attacks than Inception nets at least in the above cases, perhaps because the adversarial examples are generated from ResNet50 and are therefore more destructive to modular structures.
\begin{table}[!ht]
  \centering
  \captionsetup{width=0.5\textwidth}
  \caption{Robustness tests against noises and adversarial attacks on original and pruned Inception nets}
  \begin{tabular}{ccccc}
    \hline
     \multirow{2}{*}{Noise \& Acc Dif} 
    & \multicolumn{2}{c}{CIFAR100} & \multicolumn{2}{c}{Adience}\\
    \hhline{~----}
    & Original & Pruned & Original & Pruned\\
     \hhline{-----}
    Gaussian & -2.5\% & -2.0\% & -0.5\% & -0.1\%\\
    Poisson & -0.1\% & 0.0\% & -0.3\% & 0.0\%\\
    Speckle & -3.7\% & -3.1\% & -1.5\% & -1.0\%\\
    FGSM Attack & -8.1\% & -7.4\% & -0.4\% & -0.4\%\\
    Newton Attack & -6.1\% & -3.9\% & -4.5\% & -1.7\%\\
    \hline
  \end{tabular}
  \vspace{-0.1in}
  \begin{flushleft}
  \footnotesize{Note: for Gaussian noise, $stddev=5$. Speckle noise strength is 0.05. FGSM Attack: Fast Gradient Signed Method~\citep{goodfellow2014}. Newton Attack: Newton Fool Attack~\citep{jang2017}. For fair comparison, adversarial examples are generated against a third ResNet50 model trained with the same data.}
  \end{flushleft}
  \label{tab:noisegoogle}
\end{table}
\begin{table}[!ht]
  \centering
  \captionsetup{width=0.5\textwidth}
  \caption{Robustness tests against noises and adversarial attacks on original and pruned VGG16 nets}
  \begin{tabular}{ccccc}
    \hline
     \multirow{2}{*}{Noise \& Acc Dif} 
    & \multicolumn{2}{c}{LFWA-G} & \multicolumn{2}{c}{LFWA-S} \\
    \hhline{~----}
    & Original & Pruned & Original & Pruned \\
     \hhline{-----}
    Gaussian & -5.2\% & -4.2\% & -1.4\% & -1.2\%\\
    Poisson & 0.0\% & 0.0\% & 0.0\% & 0.0\%\\
    Speckle & -0.5\% & -0.2\% & -0.2\% & 0.0\%\\
    FGSM Attack & 0.0\% & 0.0\% & -0.1\% & 0.0\%\\
    Newton Attack & -0.2\% & -0.1\% & -3.1\% & -2.5\%\\
    \hline
  \end{tabular}
  \vspace{-0.1in}
  \begin{flushleft}
  \footnotesize{Note: for Gaussian noise, $stddev=5$. Speckle noise strength is 0.05. FGSM Attack: Fast Gradient Signed Method~\citep{goodfellow2014}. Newton Attack: Newton Fool Attack~\citep{jang2017}. For fair comparison, adversarial examples are generated against a third ResNet50 model trained with the same data.}
  \end{flushleft}
  \label{tab:noisevgg}
\end{table}

\begin{figure*}[!h]
\centering 
\begin{subfigure}{0.33\linewidth}
    \includegraphics[clip, trim=0.2in 0 0 0,width=\linewidth]{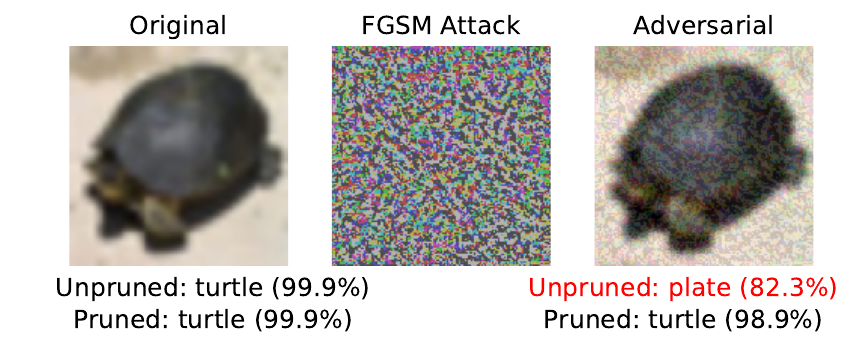}
\end{subfigure}
\begin{subfigure}{0.33\linewidth}
    \includegraphics[clip, trim=0.2in 0 0 0,width=\linewidth]{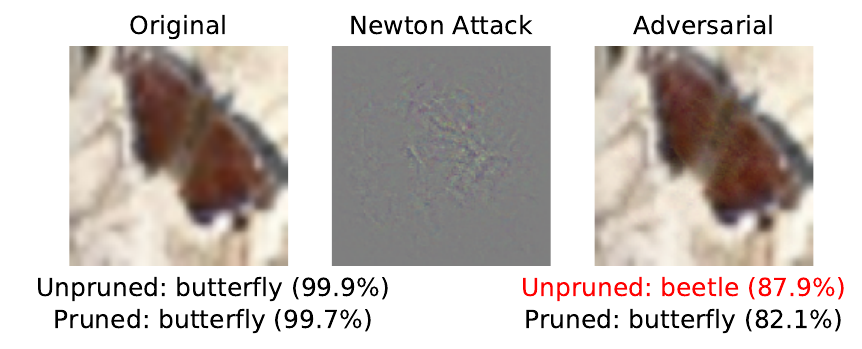}
\end{subfigure}
\begin{subfigure}{0.33\linewidth}
    \includegraphics[clip, trim=0.2in 0 0 0,width=\linewidth]{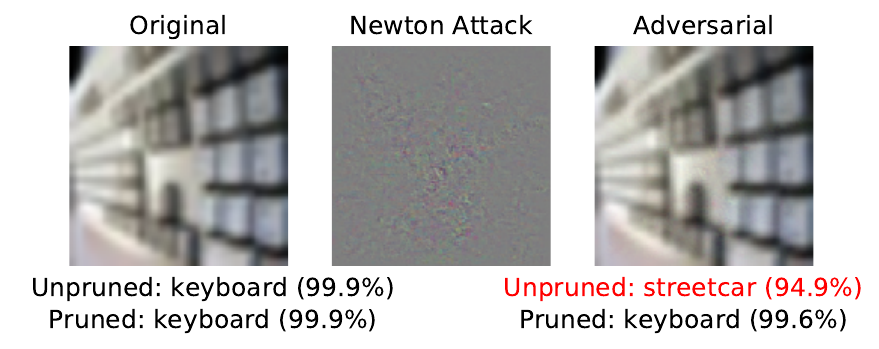}
\end{subfigure}
\newline
\begin{subfigure}{0.33\linewidth}
    \includegraphics[clip, trim=0.2in 0 0 0,width=\linewidth]{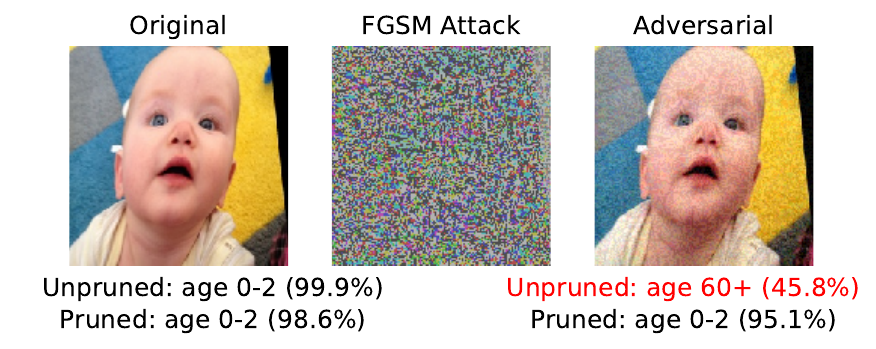}
\end{subfigure}
\begin{subfigure}{0.33\linewidth}
    \includegraphics[clip, trim=0.2in 0 0 0,width=\linewidth]{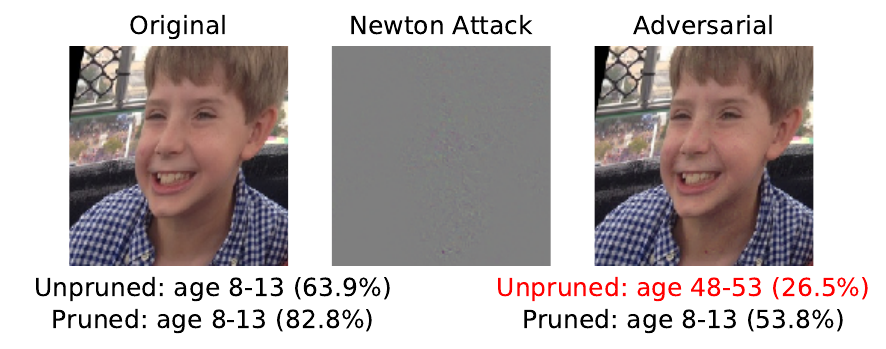}
\end{subfigure}
\begin{subfigure}{0.33\linewidth}
    \includegraphics[clip, trim=0.2in 0 0 0,width=\linewidth]{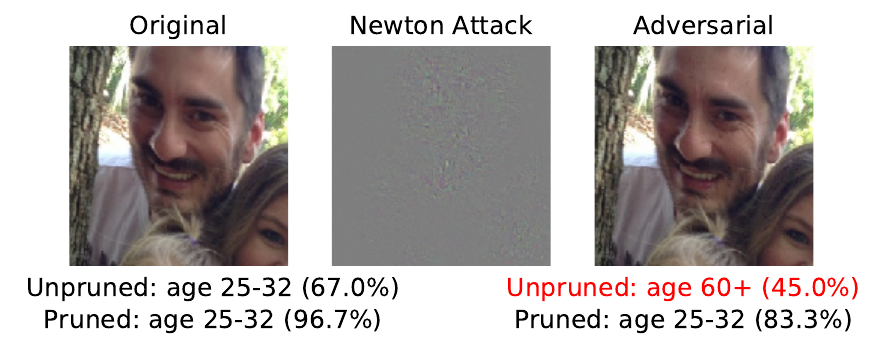}
\end{subfigure}
\newline
\begin{subfigure}{0.33\linewidth}
    \includegraphics[clip, trim=0.2in 0 0 0,width=\linewidth]{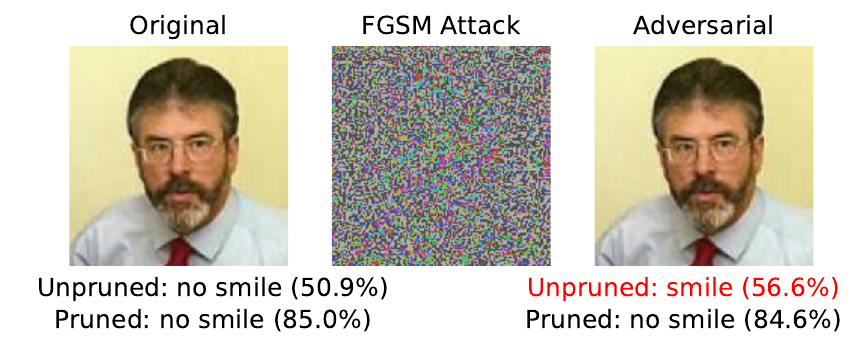} 
\end{subfigure}
\begin{subfigure}{0.33\linewidth}
    \includegraphics[clip, trim=0.2in 0 0 0,width=\linewidth]{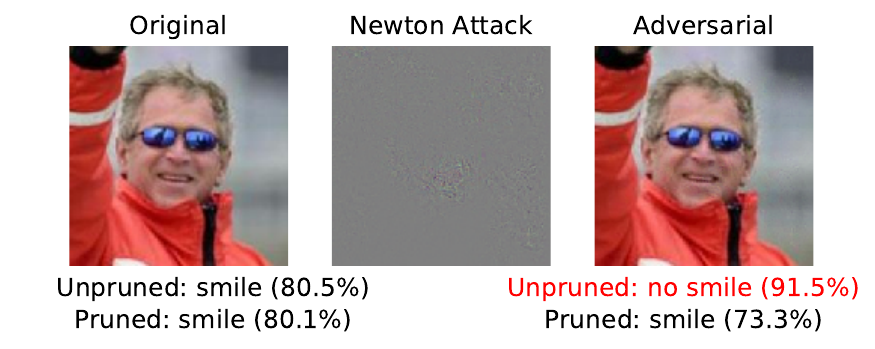} 
\end{subfigure}
\begin{subfigure}{0.33\linewidth}
    \includegraphics[clip, trim=0.2in 0 0 0,width=\linewidth]{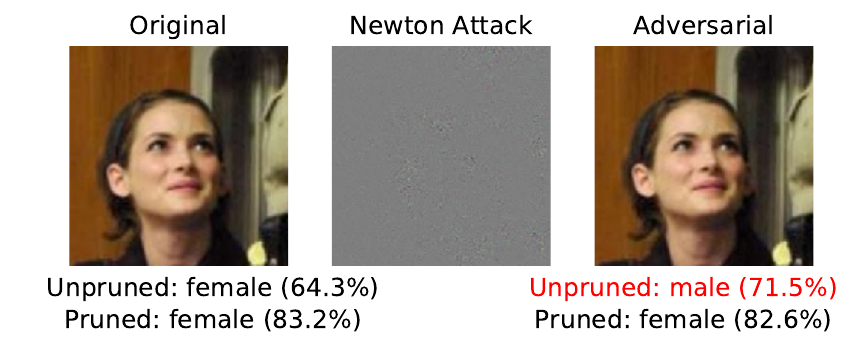}
\end{subfigure}
\caption{Example adversarial attacks that have successfully fooled the original unpruned net, but not our pruned one.}
\label{fig:adversarialdemo}
\end{figure*}

\begin{figure*}[!h]
\centering 
\begin{subfigure}{0.33\linewidth}
    \includegraphics[clip, trim=0.2in 0 0 0,width=\linewidth]{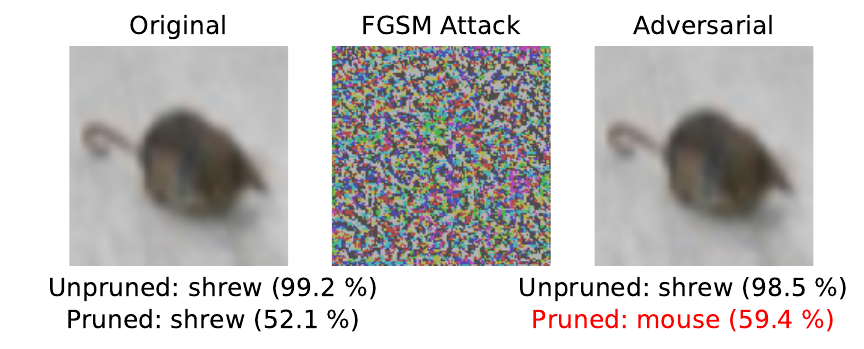}
\end{subfigure}
\begin{subfigure}{0.33\linewidth}
    \includegraphics[clip, trim=0.2in 0 0 0,width=\linewidth]{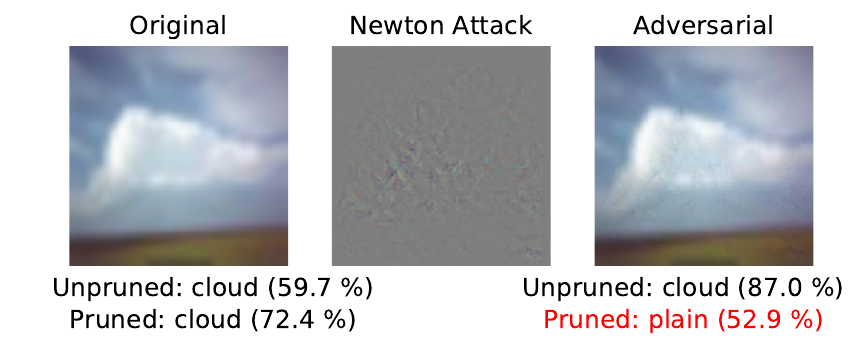}
\end{subfigure}
\begin{subfigure}{0.33\linewidth}
    \includegraphics[clip, trim=0.2in 0 0 0,width=\linewidth]{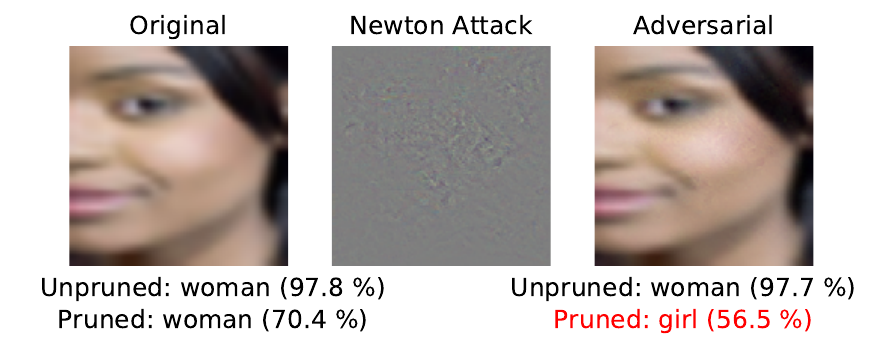}
\end{subfigure}
\newline
\begin{subfigure}{0.33\linewidth}
    \includegraphics[clip, trim=0.2in 0 0 0,width=\linewidth]{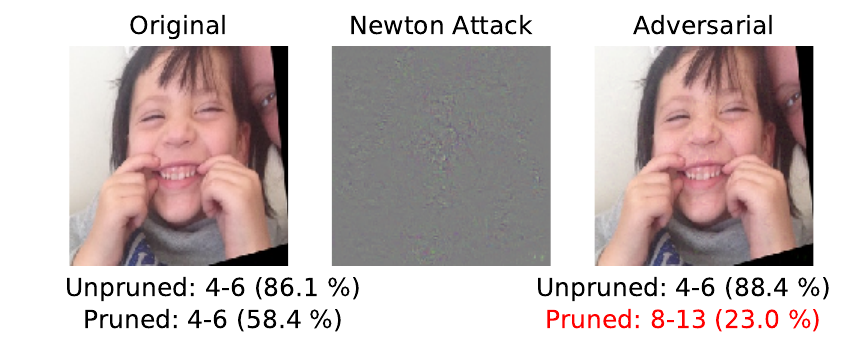}
\end{subfigure}
\begin{subfigure}{0.33\linewidth}
    \includegraphics[clip, trim=0.2in 0 0 0,width=\linewidth]{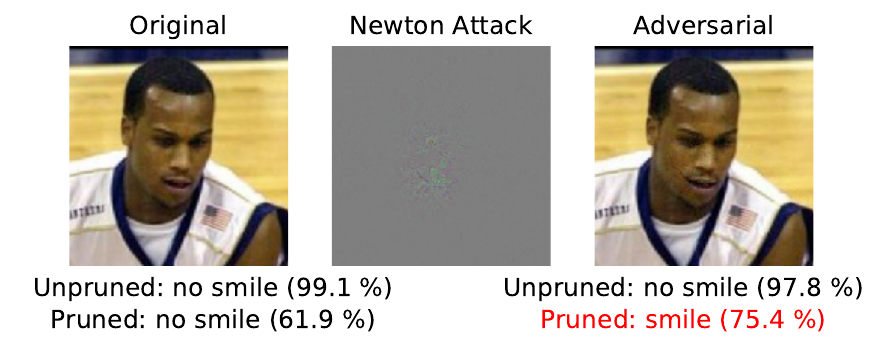}
\end{subfigure}
\begin{subfigure}{0.33\linewidth}
    \includegraphics[clip, trim=0.2in 0 0 0,width=\linewidth]{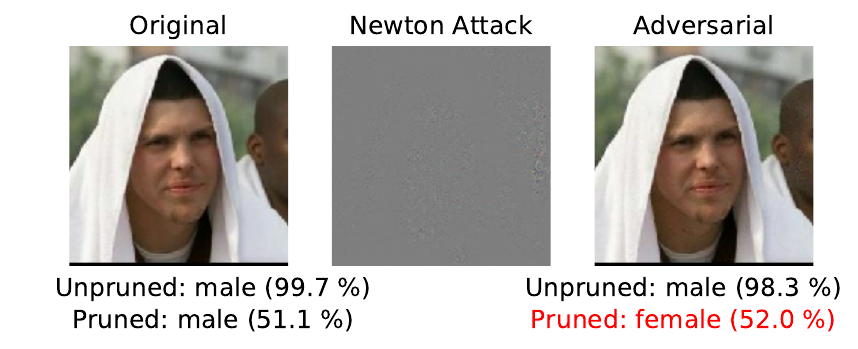}
\end{subfigure}
\caption{Example adversarial attacks that have successfully fooled our pruned net, but not the original unpruned one.}
\label{fig:adversarialantidemo}
\end{figure*}

Apart from the quantitative results, Fig.~\ref{fig:adversarialdemo} illustrates some examples where the adversarial attack fooled the original unpruned net but not our pruned one, while Fig.~\ref{fig:adversarialantidemo} shows some opposite scenarios where our pruned model failed but not the unpruned original model. The first kind of scenarios are more common across all four tasks. The examples here are randomly selected.
From the results, we can see that a small perturbation in the pixel space could make a model believe in something different. Compared to the failed cases of the pruned models in Fig.~\ref{fig:adversarialantidemo}, the fooled unpruned models in Fig.~\ref{fig:adversarialdemo} were usually very confident about their wrong predictions. The scenarios where our pruned models failed are usually ones where the pruned model was not very certain compared to the unpruned model even on the clean test data (some representative examples in our experiments are girl vs woman, house vs castle, oak tree vs forest).
Also, the nudges causing the pruned models to fail are usually more intuitive than those failed the unpruned models in Fig.~\ref{fig:adversarialdemo}. For example, while it is not directly understandable how the attacks reverted the original model's predictions about smile/no smile (the two bottom left cases in Fig.~\ref{fig:adversarialdemo}), we can see that the attack in the middle of the bottom row in Fig.~\ref{fig:adversarialantidemo} attempted to literally lift up the mouth corner into a smile (best viewed when zoomed in). Such robustness is critical. It would be disastrous if a self-driving car is easily fooled by `random noises' to believe a red light to be green.
Both of the above observations are related to the fact that large network models remember more details than the pruned ones, thus can be more confident in prediction (either correct or wrong), but sensitive to intricate data fluctuation. On the other hand, to fool a compact model pruned according to task utility, the attack has to focus on remaining task-desirable dimensions since not many irrelevant, usually easily-fooled, loophole dimensions are available.


In addition to transfer-based blackbox attacking, we have also investigated two decision-based blackbox attacks to test the unpruned and pruned models' robustness for the ImageNet case. 
Since no labelled ImageNet test set is publicly available, the experiments are carried out on the validation set.
The two decision-based attacks are SpatialAttack~\citep{engstrom2017} and ContrastReductionAttack~\citep{rauber2017}. Compared to transfer-based attacks, decision-based attacks are usually more successful given that the number of queries to the model being attacked is large enough.
Table~\ref{tab:blackboximagenet} demonstrates the success rates of the two attacks on our original and pruned models. Here, attack success rate is defined as the percentage of samples that are misclassified due to the attack. Samples that were misclassified before the attack do not count.

\begin{table}[!h]
  \centering
  \captionsetup{width=0.48\textwidth}
  \caption{Success rates of two decision-based blackbox attacks on the original and pruned models of the ImageNet case. Lower percentage means more robustness.}
  \begin{tabular}{ccc}
    \hline
     Attack \& Success rate 
    & Original Model & Pruned Model \\
     \hhline{---}
    SpatialAttack & 42.1\% & 41.4\% \\
    ContrastReductionAttack & 100.0\% & 99.9\% \\
    \hline
  \end{tabular}
  \begin{flushleft}
  \footnotesize{Note: for ContrastReductionAttack, $epsilons=3$. The attack is very successful. It fails only in 1 case when attacking the original model and fails in 23 cases when attacking the pruned model. Details of the two attacks can be found in~\citep{engstrom2017,rauber2017}.}
  \end{flushleft}
  \label{tab:blackboximagenet}
\end{table}

As we can see from Table~\ref{tab:blackboximagenet}, the SpatialAttack achieves a higher success rate on the original model than on the pruned one. The ContrastReductionAttack succeeds in fooling the two models for almost all images. The original model withstands the attack for only 1 image while our pruned model successfully defends the attack on 23 images.
The results again show that our pruning does not hurt the model robustness.

Although the results obtained in this subsection are promising for the scenarios we have investigated, more attacking strategies/algorithms and use cases need to be analyzed before drawing a more general conclusion. We defer these to future work.


\subsection{Accuracy v.s. Final Latent Space Neurons Selected}\label{sec:topneuronnum}
\begin{figure*}
\centering 
\begin{subfigure}{0.185\linewidth}
    \includegraphics[width=\linewidth,height=0.1\textheight]{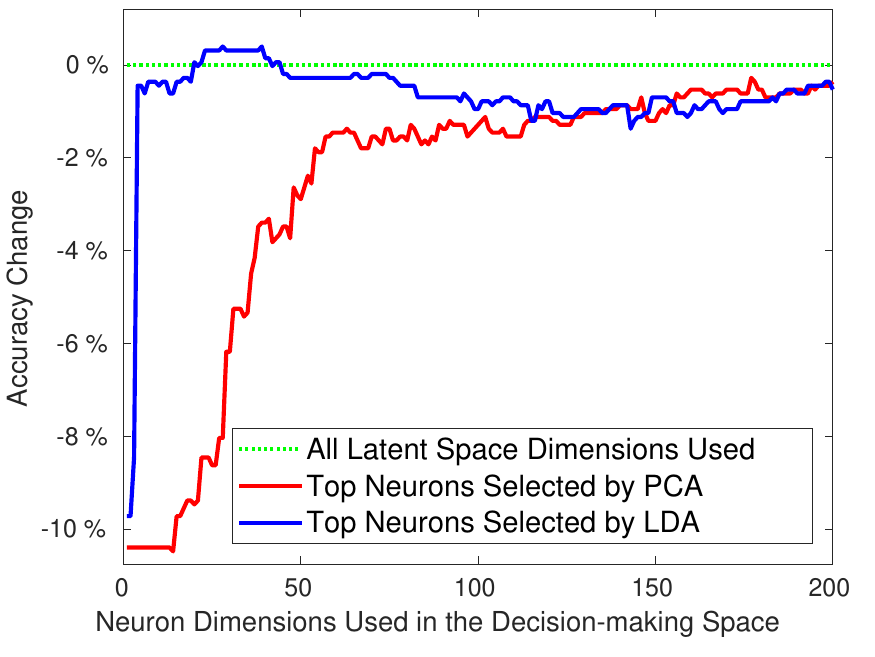}
    \caption{Gender, VGG-16}
     \label{fig:ldaneuronmale}
\end{subfigure}
~
\begin{subfigure}{0.185\linewidth}
    \includegraphics[width=\linewidth,height=0.1\textheight]{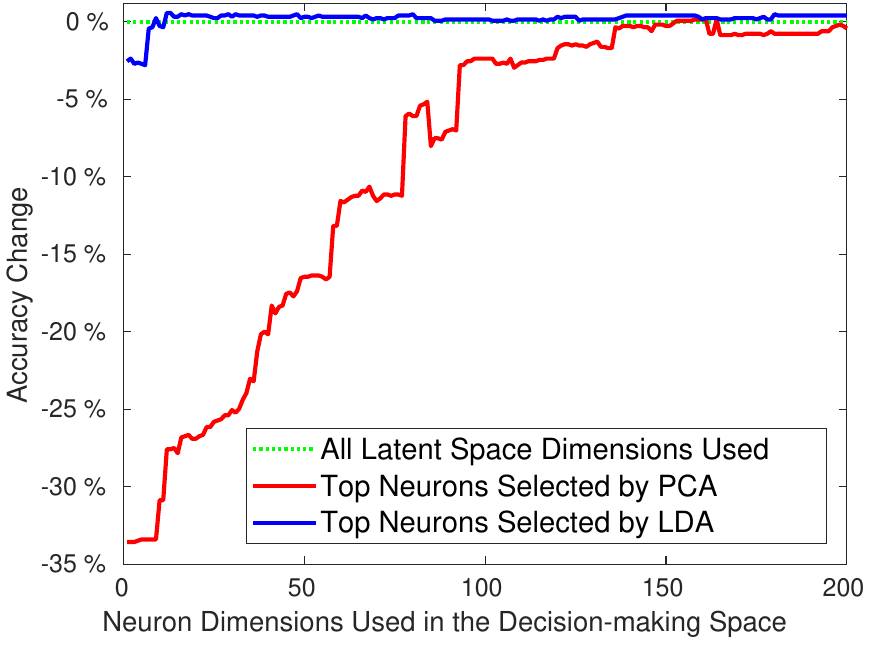}
    \caption{Smile, VGG-16}
    \label{fig:ldaneuronsmile}
\end{subfigure}
~
\begin{subfigure}{0.185\linewidth}
    \includegraphics[width=\linewidth,height=0.1\textheight]{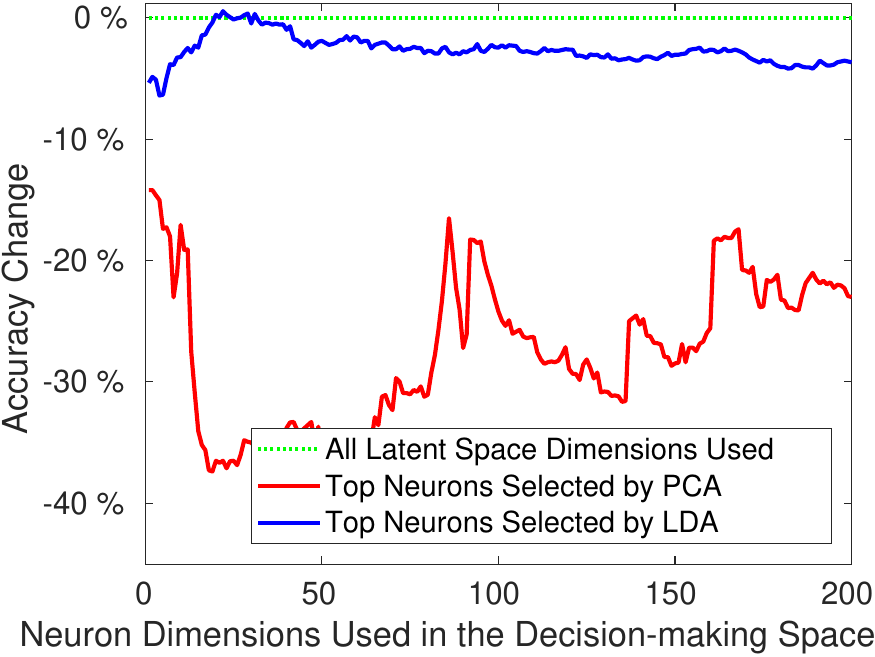}
    \caption{Age, Inception}
    \label{fig:ldaneuronage}
\end{subfigure}
~
\begin{subfigure}{0.185\linewidth}
    \includegraphics[width=\linewidth,height=0.1\textheight]{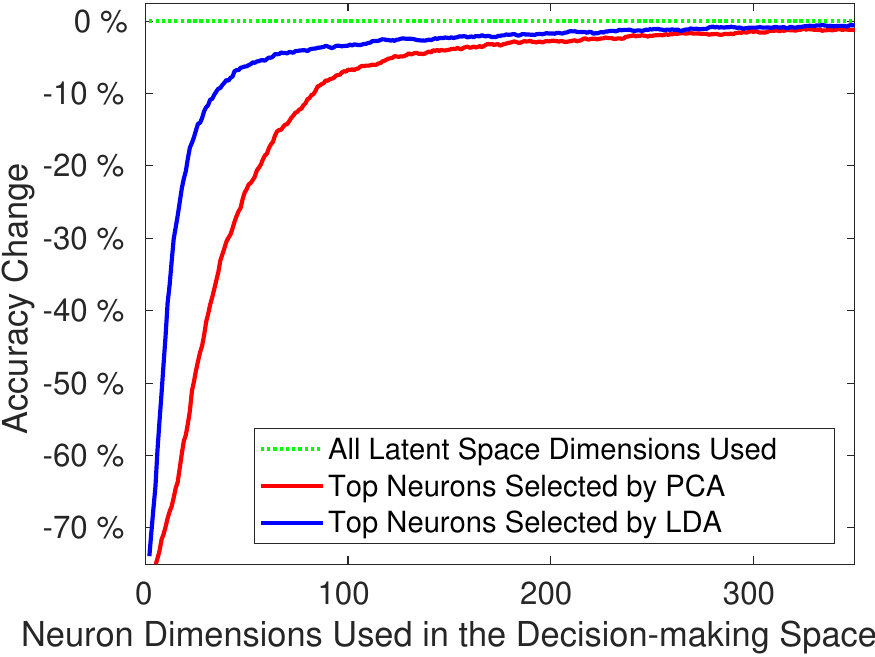}
    \caption{CIFAR100, Inception}
    \label{fig:ldaneuronsmile}
\end{subfigure}
~ 
\begin{subfigure}{0.185\linewidth}
    \includegraphics[width=\linewidth,height=0.1\textheight]{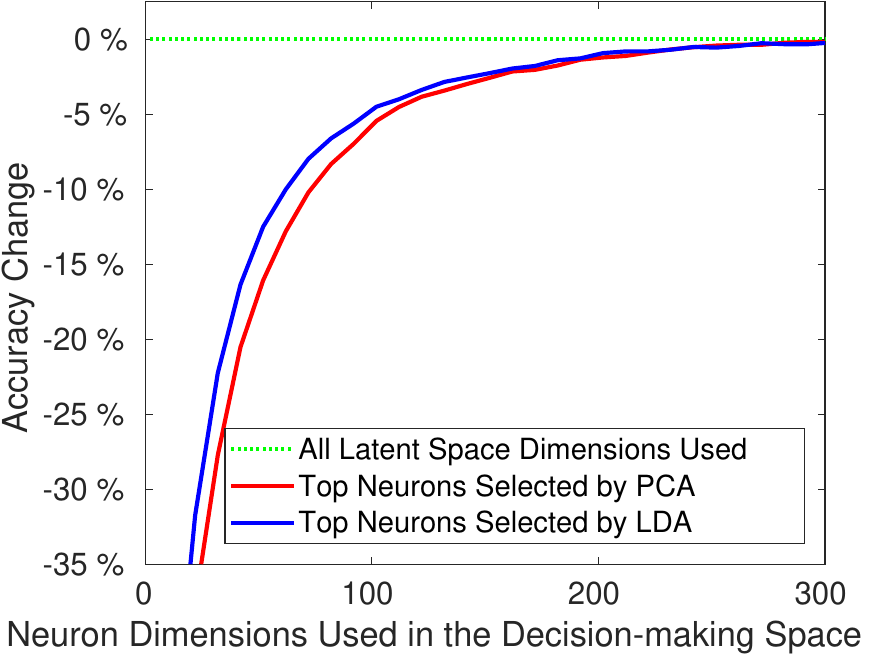}
    \caption{ImageNet, Inception}
    \label{fig:ldaneuronage}
\end{subfigure}
\caption{Accuracy change vs. number of neuron dimensions selected in decision-making space. Green dashed line indicates the accuracy when all neuron dimensions are used. Blue and red lines represent employing top neurons selected by LDA and PCA, respectively. For unpruned VGG-16 and Inception, there are respectively 4096 and 1024 neuron dimensions in the final latent space. Please note the different scales of the axes (both horizontal and vertical).}
\label{fig:LDANeuronAccuracy}
\end{figure*}
To demonstrate LDA's effectiveness in selecting final latent space neuron dimensions, we show in Fig.~\ref{fig:LDANeuronAccuracy} the relationship between accuracy change and the number of neuron dimensions preserved in the decision-making space of the base model. The top $k$ neuron dimensions with highest scores are used for the final prediction. PCA-based selection is included as comparison.
As we can see, out of thousands of latent space neuron dimensions (4096 for VGG-16, 1024 for Inception), only a small subset is capable of achieving accuracy comparable to using all dimensions. For this reason, only a subset of $k$ is plotted.
Compared to PCA, LDA performs better in all the cases. The reason is that as we increase the number of neuron dimensions, LDA is able to approximate the final class separation better and better. In contrast, PCA only explains label-blind data variation, which is not necessarily aligned with the true discriminating power. In the example of facial age recognition, people faces may vary in ethnicity, eye shape, hair style, skin color, and so on. Unlike PCA that pays attention to all such high variations, LDA picks age-related changes such as wrinkles and folds that help maximize age group separation.
When the number of neurons preserved increases, the gap between LDA and PCA becomes smaller. The gap narrows relatively fast for challenging datasets (e.g. ImageNet and CIFAR100). The reason is that, for challenging datasets, more latent space neuron dimensions are needed to maintain satisfactory accuracy. The more useful latent neuron dimensions there are, the higher the chance a useful dimension is selected even using a less accurate strategy like PCA or random sampling. It is especially true when the number of preserved neuron dimensions is large (with all neurons being selected as the extreme case). This explains the fast accuracy gain for PCA in such cases.

Apart from the above two, ICA is another linear dimension reduction technique. It could minimize dependence in the latent space before utility tracing. This has an effect of condensing information flow and reducing redundancy. Thus, `Deep ICA' can be used in unsupervised applications like auto-encoder structure design, efficient image retrieval, and image reconstruction. However, as a label-blind approach, it cannot learn class separation from groundtruth labels. The same is true to other unsupervised methods, including non-linear MDS, ISOMAP, and LLE. We do not investigate such non-linear techniques here also because over-parameterized neural networks can potentially learn the non-linearity so that we do not need extra assumptions and handcrafting. 
\cite{liu2018supervised} demonstrate the superiority of deep nets to MDS, ISOMAP, and LLE for dimensionality reduction on several benchmark datasets.
Furthermore, there are many noisy and interfering dimensions irrelevant to class separation in the original latent space. Dimension reduction preserving distance/topology in such an over-dimensioned space can possibly preserve much irrelevant information.
For example, MDS tries to preserve pairwise between-sample distances (e.g. straight-line Euclidean distance). Such distances can be easily distorted by useless dimensions~\citep{peterfreund2018multidimensional}.
Isomap preserves geodesic distance, the computation of which is also sensitive to noisy data~\citep{lee2005nonlinear}. Similarly, noise is also detrimental to LLE~\citep{chen2011locally}.

\subsection{Ablation Study of data amount for cross-layer pruning}

As mentioned previously, our pruning is data dependent. When reconstructing final class separation dependencies for pruning, we need to apply deconvolution with respect to all training images over the layers. This can be done easily for small datasets. However, for large datasets like ImageNet, this is time-consuming with limited computation resources. In this section, we explore the possibility of approximating cross-layer deep LDA utilities using only a subset of training data, and analyze its influence on final accuracy for the ImageNet case.
To this end, we choose one reference model during our pruning on ImageNet. The reference model chosen is the smallest one pruned using all training data that maintains comparable accuracy to the original (loss $<$ 1\%).

\begin{figure}[!htp]
\centering
\includegraphics[width=0.65\linewidth]{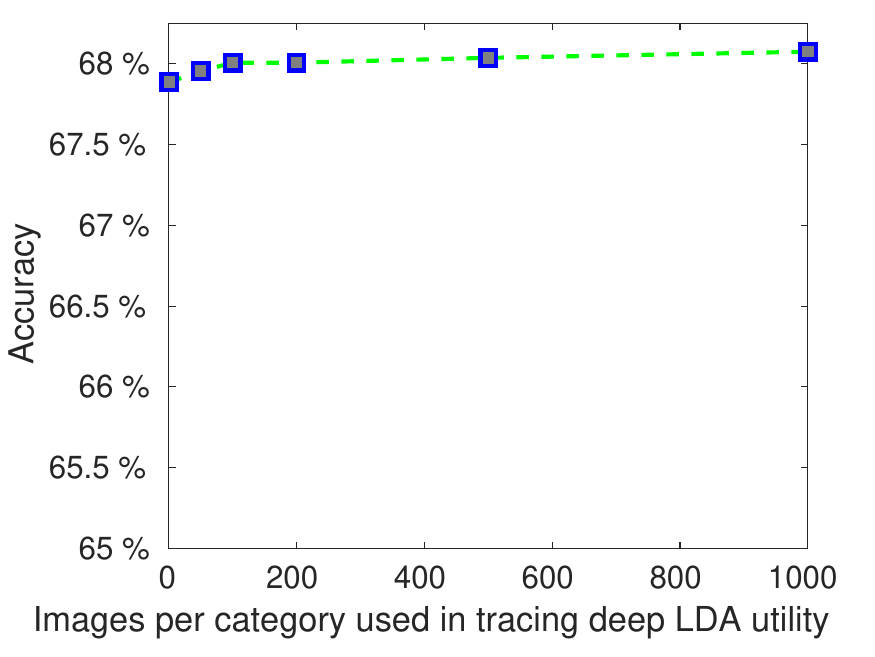}
\caption{Accuracy v.s. training images used per-category during pruning-time utility tracing. For example, 200 in the horizontal axis indicates 200 images from each category or 200,000 images in total. There are some categories with fewer than 1,000 images in ImageNet. In cases where a category runs out of images, other category images are randomly selected.}
\label{fig:tracingimgnum}
\end{figure}

We use the full training set to re-train and compute class separation utility at the top. When calculating utility dependency over the layers for pruning, we randomly select a same number of images from each category.
For certain image number selected for utility tracing, we report accuracy of the derived model after pruning and retraining. By controlling the threshold on utility scores, all resulted models in comparison here have the same parameter complexity. Figure~\ref{fig:tracingimgnum} demonstrates the results.

As we can see from Figure~\ref{fig:tracingimgnum}, accuracy is robust to image number change for pruning-time dependency tracing. With the increase of image number, accuracy only increases slightly. 1,000 images or 1 image per category can already lead to a good accuracy. 
The reason is that we trace utility only from most discriminative decision-making dimensions, so various cross-layer noisy and interfering activations are `filtered out' in the backward pass. Although images of the same category can have innumerable appearances, the essences that contribute to final class separation are limited (e.g. wrinkles and folds v.s. head pose, hair color, ethnicity for age recognition). In practice, we find that images from the same category usually lead to similar cross-layer utility dependencies (location-agnostic) except for the first few layers. This intuitively explains why performance gain saturates quickly as more training images are added.

\section{Discussion and Future Work}

In the literature, many works attempt to solve as many tasks as possible with a single generalist network. In such scenarios, large cumbersome networks are usually needed, which are impractical for situations with no or limited GPU support. A natural idea to solve this problem is to derive specialist networks~\citep{hinton2015distilling}, and each specialized network is optimized for a particular task. For a dashcam on a self-driving car, most likely, it does not need to distinguish between all the insect types and dog breeds. This paper has mainly explored the specialist path. Our pruned specialist networks have been shown to be capable of maintaining or even increasing corresponding original models' accuracy. When needed, we can flexibly form a team of expert networks specialized in different areas. This will be one of our future directions.

This paper prunes deep nets on the neuron or filter level because this directly leads to space, computation, and energy savings on general machines. That said, the proposed idea of deep discriminative dimension reduction can be applied to any, including irregular grouping of deep features, which helps select useful discriminative information at flexible granularities. Single weights and filter-based groupings are just special cases enforced by human experts. It would thus be interesting to utilize learned task-discriminative information in feature grouping/decomposition. Compared to weight sharing using conv filters, deep dimension reduction at task-desirable granularities would provide an alternative way to reducing parameter complexity which could also preserve large-scale spatial information contributing to final utility.

Another interesting direction would be to derive task-optimal architectures in a more proactive way by pushing useful deep discriminants into alignment with a condensed subset of neurons (or other easily-pruned substructures) before deconv based deep feature decomposition and reduction. In a concurrent work, we achieve this by including deep LDA utility and covariance penalty simultaneously in the objective function~\citep{tian2020deep}. That said, compared to the simple pruning method presented here, proactive eigen-decomposition and training can be computationally expensive and numerically unstable.

\section{Conclusion} \label{sec:conclusion}

This paper proposes a task-dependent end-to-end pruning approach with a deep LDA utility that captures both final class separation and its holistic cross-layer dependency. This is different from approaches that are blind or pay no direct attention to task discriminative power and those with local (individual weights or within 1-2 layers) utility measures. The proposed approach is able to prune convolutional, fully connected, modular, and hybrid deep structures and it is useful for designing deep models by finding both the desired types of filters and the number for each kind. Compared to fixed nets, the method offers a range of models that are adapted for the inference task in question. On datasets of general objects and domain specific tasks (ImageNet, CIFAR100, LFWA and Adience), the approach achieves better performance and greater complexity reductions than competing methods and models.
Moreover, the method is shown to be capable of generating compact models that are more robust to adversarial attacks and noises than the original unpruned model. The approach's global awareness of task discriminating power, high pruning rates, and its resulting models' robustness offer a great potential for installation of deep nets on mobile devices in many real-world applications.

\section*{Acknowledgments}
This research was enabled in part by support provided by Calcul Québec (\url{www.calculquebec.ca}) and Compute Canada (\url{www.computecanada.ca}). Also, the authors would like to gratefully acknowledge the support from the Natural Sciences and Engineering Research Council of Canada.

\bibliographystyle{model2-names}
\bibliography{ycviu-template-with-authorship}

\end{document}